# How do features act in an individual prediction? Separating standalone and interaction effects with a case-based taxonomy

Sejong Oh

Department of Software Science, Dankook University, Yongin, Republic of Korea

sejongoh@dankook.ac.kr

## Abstract

**Background.** Local explanation methods for tabular models, such as SHAP and LIME, summarize each feature with a single attribution score. Interaction-specific indices, such as SHAP interaction values or the Friedman H-statistic, quantify how strongly features interact. Neither tells a user *how* a feature acts in a given prediction: whether its effect is the same on its own and in context, or whether the other features amplify it, make it unnecessary, switch it off, or reverse its direction.

**Methods.** We extend case-based feature importance (CBFI), originally a global method, to individual predictions. For each feature and each background sample, the model is evaluated in four hybrid states, which give a standalone effect, an in-context effect, the effect of the remaining features, and the interaction effect as the difference between the first two. Their averages decompose the in-context importance of a feature into standalone and interaction components. A case taxonomy then assigns each feature of an instance one of six labels: inactive, standalone, amplified, redundant, gated or reversed. The labels are computed from bootstrapped means with a practical-relevance threshold, and they correspond to the groups of the original CBFI. We evaluate the recovery of known interactions against k-SII, TreeSHAP interaction values and the H-statistic on synthetic functions, distinguishing interactions in the data-generating process from interactions learned by fitted models. We further assess mechanism identification, label robustness, stability, cost, scalability and sampling sensitivity, and study eight real tabular datasets with six model families.

**Results.** On the true functions, the decomposition recovered interactions exactly, with no false interactions and a correlation of 0.99 with the analytic local interaction. On fitted models, pair recovery was close to but slightly below that of dedicated indices (local precision@k 0.86–0.91 versus 0.91–0.95). The distinctive output of the method was the mechanism label. Interaction values alone could not distinguish substitution from gating (accuracy 0), whereas the taxonomy identified both without knowing the interacting partner (0.74–0.76). Applying the same taxonomy to Shapley interaction values, using the attribution of the remaining features, performed comparably (overall 0.88–0.91 versus 0.86–0.88). On fitted models the labels agreed with the generating logic in 95% of cases, and on real data 45–57% of feature pairs with nearly identical SHAP values received different labels. In a medical-insurance model, the method showed that smoking has a large standalone effect whose interaction component changes sign at a body-mass index of 30; its in-context importance agreed with a counterfactual check ($r = 0.96$). A two-stage screening procedure kept the cost roughly linear in the number of features.

**Conclusions.** Localized CBFI does not detect interactions more accurately than existing indices. What it adds to the explanation of an individual prediction is a decomposition into standalone and interaction effects and a label stating how each feature acts, information that attribution scores and interaction values do not provide directly. The taxonomy is not tied to our estimator and can also be layered on Shapley interaction indices.

## Introduction

Machine learning models for tabular data are increasingly used in settings where decisions must be justified, for example in medicine, insurance and finance (Rudin, 2019; Adadi & Berrada, 2018). Post-hoc local explanation methods address this need by explaining individual predictions. The most widely used of them, SHAP (Lundberg & Lee, 2017) and LIME (Ribeiro, Singh & Guestrin, 2016), express a prediction as a sum of feature contributions. Such additive attributions are easy to read, but a single score per feature cannot show whether the feature acts on its own or only in combination with others. The contribution of an interaction is distributed over the participating features, so two features with similar scores may play entirely different roles in the model (Molnar, 2025; Bordt & von Luxburg, 2023).

A substantial body of work has therefore been devoted to quantifying interactions explicitly. At the global level, Friedman's H-statistic measures the share of the variance of partial dependence that is due to interaction (Friedman & Popescu, 2008), and functional decompositions separate main effects from interactions (Hooker, 2007; Herbinger et al., 2024). At the local level, SHAP interaction values can be computed exactly for tree ensembles (Lundberg et al., 2020), and a family of Shapley interaction indices, including the Shapley–Taylor index (Sundararajan, Dhamdhere & Agarwal, 2020), Faith-Shap (Tsai, Yeh & Ravikumar, 2023) and k-SII (Bordt & von Luxburg, 2023), is now available in efficient implementations for any model and any order (Muschalik et al., 2024a; Muschalik et al., 2024b). For neural networks, dedicated detectors search for interacting feature sets in the trained model (Tsang, Cheng & Liu, 2018; Tsang, Rambhatla & Liu, 2020). All of these methods answer the question of *which* features interact and *how strongly*.

For a user who must understand a single prediction, a second question is often just as important: *how* does a feature act in this prediction? Its effect may be the same whether or not the other features are present. The other features may amplify it, as in a synergistic AND-type relation, or they may already carry the same information and make it unnecessary, as in an OR-type relation. An unmet condition on another feature may switch its effect off, and in some cases the context even reverses the direction of the effect. These situations are qualitatively different, yet an interaction value does not distinguish them. For example, an OR function evaluated where both inputs are active and an AND function evaluated where one input is inactive yield the same pairwise Shapley interaction value, although the underlying logic is opposite.

Case-based feature importance (CBFI) was proposed as a global importance measure that classifies instances into four groups: those for which a feature is sufficient on its own (G1), those for which the remaining features suffice (G2), those for which either suffices (G3), and those for which only their combination succeeds (G4) (Oh, 2022). This view of sufficiency and joint necessity is close to the qualitative question raised above, but it operates on a whole dataset and on the correctness of classification. In this study, we localize CBFI to individual predictions and turn it into a quantitative decomposition with a qualitative interpretation.

The contributions of this study are as follows:

1. We define, for every feature of an instance, a standalone effect, an in-context effect and an interaction effect by comparing four hybrid states of the input against background samples. The in-context importance is exactly the sum of the standalone and interaction components, and the interaction component vanishes for features that enter the model additively.
2. We introduce a case taxonomy that labels each feature of an instance as inactive, standalone, amplified, redundant, gated or reversed. The labels are connected to the groups of the original CBFI and are made robust for fitted models through bootstrapped means and an explicit practical-relevance threshold. The taxonomy requires three quantities, namely the standalone, in-context and remaining-feature effects; localized CBFI computes them directly, and they can also be approximated from Shapley-type attributions.

3. We evaluate the method against known ground truth and against interaction-specific methods (k-SII, TreeSHAP interaction values and the H-statistic) under identical conditions, separating interactions in the data-generating process from interactions learned by the model, and we report where the method is less accurate or more costly than the existing alternatives.
4. Using eight tabular datasets and six model families, we show what the decomposition and the labels add to the explanation of individual predictions, propose a two-stage screening procedure for high-dimensional data, and quantify how much information the labels add to an attribution score.

We do not claim that localized CBFI detects interactions more accurately than dedicated interaction indices. Its purpose is to add information that attribution scores do not convey: alongside the importance of each feature, it states how much of that importance depends on the context and in what way. As our experiments show, features with nearly equal attribution scores often differ in exactly this respect.

## Related work

### *Feature attribution and effect visualization*

Additive attribution methods assign each feature a contribution to a prediction. SHAP derives the contributions from the Shapley value of a cooperative game (Lundberg & Lee, 2017), and LIME fits a local linear surrogate model around the instance (Ribeiro, Singh & Guestrin, 2016). Permutation feature importance measures the increase in loss when a feature is permuted, either globally or for whole classes of models (Breiman, 2001; Fisher, Rudin & Dominici, 2019). Covert, Lundberg & Lee (2021) showed that many of these methods can be unified as removal-based explanations, which differ in how features are removed and how the resulting effect is summarized. Visual methods take a complementary approach and describe how the prediction changes with a feature, through partial dependence plots (Friedman, 2001), individual conditional expectation curves (Goldstein et al., 2015) or accumulated local effects (Apley & Zhu, 2020).

Rule-based local explanations follow a different route. Anchors (Ribeiro, Singh & Guestrin, 2018) and LORE (Guidotti et al., 2018) give sufficient conditions for a prediction, and Contextual Importance and Utility relates the importance of a feature to the range of outputs it can produce in the current context (Främling, 2020). Reviews of the field discuss these families and their evaluation (Adadi & Berrada, 2018; Ortigossa, Gonçalves & Nonato, 2024; Molnar, 2025). An important practical observation is that different attribution methods frequently disagree on the same prediction, and agreement between two methods cannot by itself establish that either of them is faithful to the model (Krishna et al., 2024). The quantities used in this study are closely related to partial dependence and to local permutation effects, as shown in the Materials & Methods section. What localized CBFI adds is the explicit difference between them and its qualitative interpretation.

### *Quantifying feature interactions*

Friedman's H-statistic measures, at the global level, the fraction of the variance of a two-feature partial dependence that is not explained by the one-feature partial dependences (Friedman & Popescu, 2008). Functional ANOVA decompositions generalize this idea to dependent inputs (Hooker, 2007), and more recent work partitions the feature space into regions in which the interaction-related heterogeneity of local effects is small (Herbinger et al., 2024).

Shapley-based interaction indices form the largest group of local methods. SHAP interaction values split the Shapley value of a feature into a main effect and pairwise interactions and can be computed exactly for tree ensembles (Lundberg et al., 2020). The Shapley interaction index of cooperative game theory (Grabisch & Roubens, 1999) and its efficient variants distribute the prediction over interactions up to a chosen order; examples are the Shapley–Taylor interaction index (Sundararajan, Dhamdhere & Agarwal, 2020), Faith-Shap (Tsai, Yeh & Ravikumar, 2023), and k-SII and n-Shapley values (Bordt & von Luxburg, 2023). Bordt & von Luxburg (2023) also showed that such indices of order n recover generalized additive models with interactions up to order n. The shapiq library implements these indices for any model, together with approximation algorithms and the exact TreeSHAP-IQ algorithm for tree ensembles (Muschalik et al., 2024a; Muschalik et al., 2024b). Shapley interactions have also been applied beyond prediction, for example to explain hyperparameter optimization (Wever et al., 2026), and sparse interaction discovery has been scaled to large language models (Butler et al., 2025). For neural networks, interacting feature sets can be identified from the learned weights or by perturbation (Tsang, Cheng & Liu, 2018; Tsang, Rambhatla & Liu, 2020).

These approaches quantify the strength and sign of interactions. They do not, however, tell the user whether a negative interaction reflects two features substituting for each other or one feature being switched off by another. A closely related idea is the decomposition of feature relations into synergy, redundancy and independence, which has been developed in information theory (Williams & Beer, 2010) and for global SHAP-based model explanations (Ittner et al., 2021). Our taxonomy differs from these in three respects: it operates on individual predictions, it separates redundancy (substitution) from gating, and it identifies a reversal of the direction of a feature's effect, all without requiring the user to know which feature is the interacting partner.

### *Reference distributions and dependent features*

Every removal-based method must decide what a "removed" feature is replaced with. Marginal, or interventional, replacement can create unrealistic combinations when features are correlated and thereby force the model to extrapolate (Hooker, Mentch & Zhou, 2021). Conditional replacement preserves the joint distribution but changes the question being answered (Aas, Jullum & Løland, 2021). Whether an explanation should be true to the model or true to the data therefore depends on its purpose (Janzing, Minorics & Blöbaum, 2020; Chen et al., 2020). In this study, we use marginal replacement by default and examine the effect of conditional replacement explicitly.

### *Research gap addressed in this study*

Existing interaction methods answer *which* features interact and *how strongly*, and existing local attributions answer *how important* each feature is. What remains unaddressed is a statement, for each prediction, of *how* each feature acts: on its own, amplified, made redundant, gated or reversed by the context. Such a statement should be obtained without enumerating and inspecting candidate pairs, and it should be validated on fitted models against known ground truth. The present study addresses this gap and quantifies its cost relative to the existing methods.

## Materials & Methods

### *Global case-based feature importance*

Global CBFI (Oh, 2022) compares, for every instance of a dataset, the prediction of a classifier under three conditions: on the original data, on data in which all features except a target feature $F_x$ are permuted ("$F_x$ alone"), and on data in which only $F_x$ is permuted ("the rest without $F_x$"). Instances that are predicted correctly on the original data are then assigned to four groups. In group G1, $F_x$ alone suffices for a correct prediction but the rest does not; in G2, the rest suffices but $F_x$ alone does not; in G3, each of them suffices; and in G4, neither suffices, but their combination does. The importance of $F_x$ is the share of G1 and G4 instances. Localized CBFI transfers this logic of sufficiency and joint necessity to a single instance and to a continuous model output.

***Notation and explained output***

Let $f$ be a trained model and $x = (x_1, \dots, x_p)$ the instance to be explained. For a set of features $S$ and a background row $z$, $(x_S, z_{-S})$ denotes the input that takes the features in $S$ from $x$ and all other features from $z$. The explained quantity $v$ is the model output. For regression, $v = f$ is the prediction; for classification, $v$ is the predicted probability of the class that the model predicts for $x$. True labels are not used, so explanations can also be computed for unlabeled instances. All features enter the explainer in their original units, and any scaling is treated as part of the model pipeline.

***Sampling of background data***

We draw $n$ background rows $z^{(1)}, \dots, z^{(n)}$ uniformly from the training data (marginal sampler). As an alternative, we implement a conditional sampler that draws from the $k$ nearest training rows of $x$, where the distance is measured on the two features most correlated with the target feature after standardization.

***Four hybrid states and per-sample effects***

For a target feature $i$ and each background row $z^{(s)}$, the model is evaluated in four states (Fig. 1A): the full instance, $v(x)$; the instance with feature $i$ replaced, $v(x_{-i}, z_i^{(s)})$; the background row with only feature $i$ taken from the instance, $v(x_i, z_{-i}^{(s)})$; and the background row itself, $v(z^{(s)})$. These four evaluations give four per-sample quantities (Fig. 1B). The in-context effect of feature $i$ is

$$b_s = v(x) - v(x_{-i}, z_i^{(s)}), \quad (1)$$

the standalone effect of feature $i$ is

$$a_s = v(x_i, z_{-i}^{(s)}) - v(z^{(s)}), \quad (2)$$

and the effect of the remaining features is

$$c_s = v(x_{-i}, z_i^{(s)}) - v(z^{(s)}). \quad (3)$$

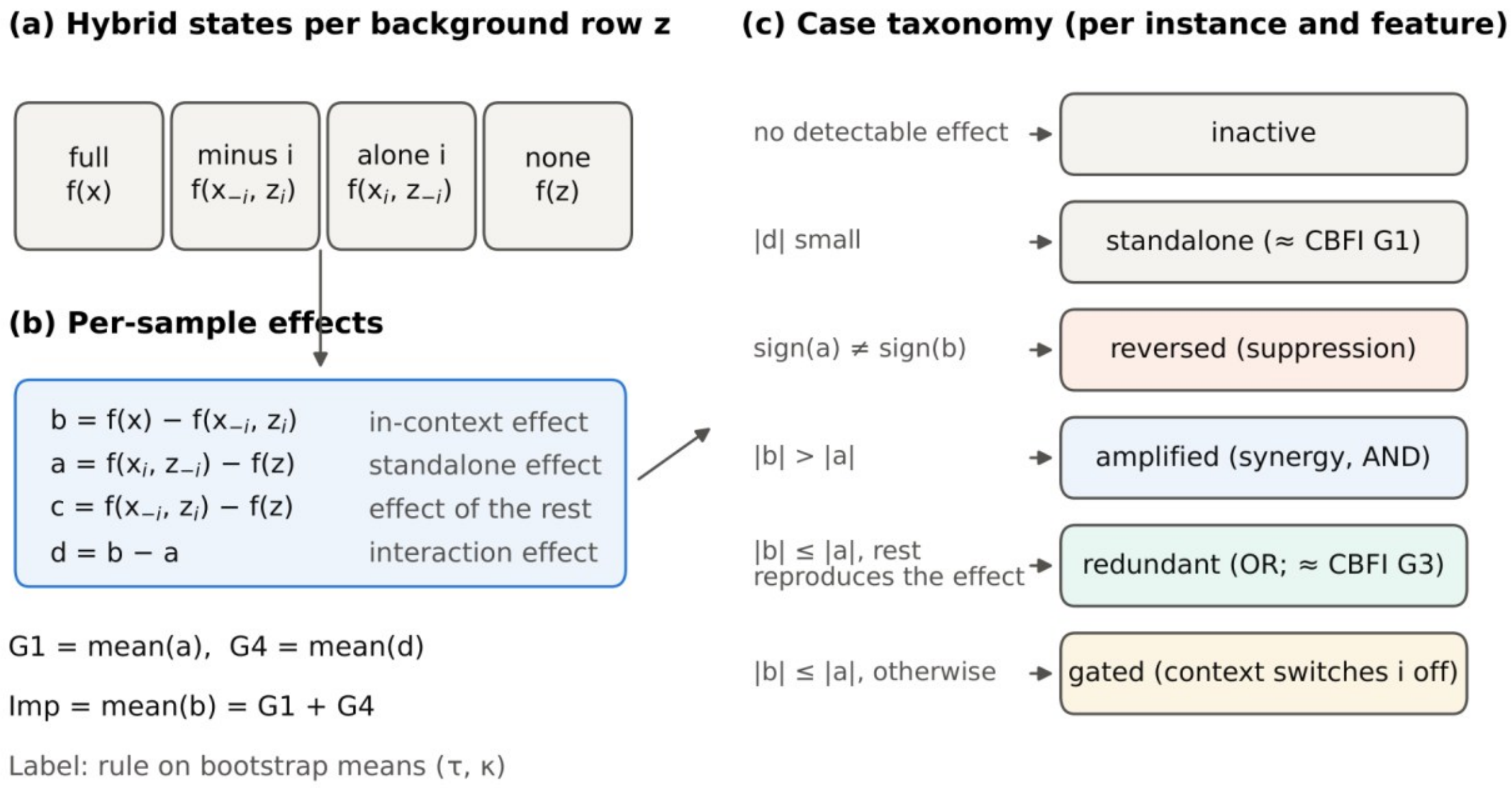

**Figure 1. Localized CBFI.** (A) For every background row z, the model is evaluated in four hybrid states built from the explained instance x and z. (B) Per-sample effects: in-context effect b, standalone effect a, effect of the remaining features c, and interaction effect d = b − a. Their means give the standalone effect G1, the interaction effect G4 and the in-context importance Imp = G1 + G4. (C) The case taxonomy assigns each feature of an instance one of six labels from the bootstrap means of a, b and c, using a practical threshold τ and a relative threshold κ. The labels correspond to the groups of the original global CBFI (standalone ≈ G1, amplified ≈ G4, redundant ≈ G3).

The interaction effect of feature $i$ is the difference between the in-context and standalone effects:

$$d_s = b_s - a_s = v(x) - v(x_{-i}, z_i^{(s)}) - v(x_i, z_{-i}^{(s)}) + v(z^{(s)}). \quad (4)$$

Eq. (4) is the second-order difference between feature $i$ and the set of all other features, evaluated against the reference $z^{(s)}$. If $f$ is additive in feature $i$, that is, $f(x) = g(x_i) + h(x_{-i})$, then $d_s = 0$ for every $s$.

***Decomposition into standalone and interaction effects***

Averaging the per-sample quantities over the background samples gives the standalone effect, the interaction effect and the in-context importance of feature $i$:

$$G_{1,i}(x) = \frac{1}{n}\sum_{s=1}^{n} a_s, \quad (5)$$

$$G_{4,i}(x) = \frac{1}{n}\sum_{s=1}^{n} d_s, \quad (6)$$

$$\mathrm{Imp}_i(x) = \frac{1}{n}\sum_{s=1}^{n} b_s = G_{1,i}(x) + G_{4,i}(x). \quad (7)$$

All three quantities are signed and expressed in the units of the output, that is, in currency for the insurance model and in probability for classification. Because positive and negative values of $d_s$ can cancel in Eq. (6), we also report the mean absolute interaction $\frac{1}{n}\sum_s |d_s|$. In the remainder of the paper, we write $G_1$, $G_4$ and Imp when the feature and the instance are clear from the context.

***A case taxonomy of feature behaviour***

The taxonomy states how feature $i$ acts at instance $x$ (Fig. 1C). It is evaluated on the means $\overline{A}$, $\overline{B}$, $\overline{C}$ and $\overline{D}$ of $a_s$, $b_s$, $c_s$ and $d_s$, using a practical threshold $\tau$, which is the smallest effect regarded as relevant, and a relative threshold $\kappa$. The rules are applied in the following order:

- **inactive** if $|\overline{A}| \leq \tau$ and $|\overline{B}| \leq \tau$: the feature has no relevant effect;
- **standalone** if $|\overline{D}| \leq \max(\tau, \kappa\max(|\overline{A}|, |\overline{B}|))$: the feature acts in the same way on its own and in context;
- **reversed** if $|\overline{A}| > \tau$, $|\overline{B}| > \tau$ and $\mathrm{sign}(\overline{A}) \neq \mathrm{sign}(\overline{B})$: the context reverses the direction of the feature's effect;
- **amplified** if $|\overline{B}| > |\overline{A}|$: the context strengthens the feature's effect (synergy, AND-type relation);
- **redundant** if $\overline{C} \cdot \mathrm{sign}(\overline{A}) \geq |\overline{D}|/2$: the context weakens the feature's effect, and the remaining features alone reproduce at least half of the lost effect (substitution, OR-type relation);
- **gated** otherwise: the context weakens or switches off the feature's effect without substituting for it, as for an unmet AND condition.

The labels correspond to the groups of the global method. The standalone label corresponds to G1 (the feature alone suffices), amplified to G4 (joint necessity) and redundant to G3 (either suffices). G2 describes the remaining features rather than $F_x$ and is represented locally by the inactive label. The gated and reversed labels have no global counterpart, because they become visible only at the level of individual instances.

Fitted models, in particular tree ensembles, are piecewise constant. Individual background samples can therefore show spurious differences between $a_s$ and $b_s$ even for features that act additively. For this reason, the rules are applied to the means rather than to the individual samples. We draw 300 bootstrap replicates of the background samples, apply the rules to each replicate, and report the modal label together with its frequency, which we call the label confidence. In addition, a feature is labelled inactive when the 95% bootstrap intervals of both $\overline{A}$ and $\overline{B}$ contain zero, and standalone when the interval of $\overline{D}$ contains zero. Unless stated otherwise, we use $\tau = 0.1 \times \mathrm{sd}(v(z))$ and $\kappa = 0.1$. The threshold $\tau$ is an interpretive parameter, namely the smallest effect a user considers relevant, and its influence is examined in the experiments. For descriptive purposes, we also record the share of background samples that fall into each case when the same rules are applied to each sample with $\tau = 0.01 \times \mathrm{sd}(v(z))$; these shares are not used for labelling.

***Pairwise interaction and two-stage screening***

To identify the partner of a feature, we compute the pairwise interaction of features $i$ and $j$ while all other features are kept at their values in $x$:

$$G_{4,ij}(x) = \frac{1}{n}\sum_{s=1}^{n} [v(x) - v(x_{-i}, z_i^{(s)}) - v(x_{-j}, z_j^{(s)}) + v(x_{-ij}, z_{ij}^{(s)})]. \qquad (8)$$

$G_{4,ij}$ is the effect of feature $i$ when feature $j$ is present minus the effect of $i$ when $j$ is removed. It is symmetric in $i$ and $j$, and it is zero for every sample when $f$ contains no term that couples $i$ and $j$. Computing Eq. (8) for all pairs requires $O(p^2 n)$ model evaluations. For data with many features, we therefore use a two-stage procedure. In the first stage, the features are ranked by their mean absolute interaction $\frac{1}{n}\sum_s |d_s|$, which needs only $O(pn)$ evaluations; in the second stage, $G_{4,ij}$ is computed only among the top $m$ features ($m = 12$ in our experiments).

***Properties of the decomposition***

The decomposition has several properties that follow directly from its definition. First, $\mathrm{Imp}_i = G_{1,i} + G_{4,i}$ holds exactly for every instance. Second, if $f$ is additive in feature $i$, then $G_{4,i} = 0$, and if $f$ has no term coupling $i$ and $j$, then $G_{4,ij} = 0$. Third, the quantities are closely related to existing ones. $G_{1,i}(x)$ equals the centered partial dependence of $f$ at $x_i$ under the chosen sampler (Friedman, 2001; Goldstein et al., 2015), and $\mathrm{Imp}_i(x)$ is a local, signed leave-one-out (permutation) effect. Eq. (8) is the interventional second-order difference evaluated in the context of $x$, and k-SII and related Shapley interaction indices average differences of the same kind over coalitions (Bordt & von Luxburg, 2023; Muschalik et al., 2024a). The decomposition is therefore not a new interaction functional; what it adds is the case taxonomy, which uses $a_s$, $b_s$ and $c_s$ jointly, and the single-feature screen.

The range of the quantities depends on the task. For classification, $a_s$, $b_s$ and $c_s$ lie in $[-1,1]$ and $d_s$ in $[-2,2]$; for regression, they are bounded by the range of $f$ over the background data. As with SHAP values, magnitudes are interpreted relative to the spread of the output, which is why $\tau$ is expressed as a fraction of $\mathrm{sd}(v(z))$.

The sign of the interaction effect also requires care. $G_{4,i} > 0$ means that the joint presence of feature $i$ and the rest of $x$ pushes the output higher than feature $i$ would on its own, and $G_{4,i} < 0$ means that it pushes the output lower. This sign is local and relative to the reference distribution. For a term $cx_ix_j$ with means $\mu_i$ and $\mu_j$, the expected per-sample interaction is $c(x_i - \mu_i)(x_j - \mu_j)$. A term with a negative coefficient can therefore produce a positive $G_4$ for instances on one side of the mean, and a "suppressor" term does not imply $G_4 < 0$. The direction-free labels (amplified, redundant, gated and reversed) are provided for this reason.

***Computational cost and implementation***

The decomposition of one feature requires $3n$ model evaluations, which are batched into single prediction calls, and explaining all features of an instance requires about $3np + n$ evaluations. With $n = 100$, the Monte Carlo variability of $G_4$ relative to $|\text{Imp}|$ is about 8%, as shown in the stability experiment. The implementation is available at https://github.com/dkumango/Local_CBFI.

## Experimental design

***Evaluation questions***

The evaluation is organized around five questions. The first question (Q1) is whether localized CBFI recovers known interacting feature pairs and how it compares with interaction-specific indices. The second (Q2) goes beyond *whether* a feature interacts and asks whether the method can tell *how* it interacts at a given instance, that is, whether the feature is amplified, redundant, gated or reversed, and how reliable these labels are on fitted models. The third (Q3) concerns the Monte Carlo variability of the estimates, the cost of an explanation and the behaviour of the method as the number of features grows. The fourth (Q4) examines how sensitive the explanations are to replacing marginal sampling with conditional sampling. The last question (Q5) asks what the decomposition reveals on real tabular data, how it relates to SHAP, LIME and a counterfactual check, and how much information the labels add to an attribution score.

***Two levels of ground truth: data-generating process and fitted model***

An interaction can exist in the data-generating process (DGP), that is, in the true function that produced the data, or only in the model fitted to the data, and the two need not coincide. We therefore run every synthetic experiment at two levels. At the DGP level, the true function itself is used as the model, so any error is due to the explanation method. At the model level, a random forest (RF), XGBoost (XGB) and a multilayer perceptron (MLP, also denoted ANN in the figures) are fitted to noisy samples ($\sigma = 0.05$). Differences from the DGP level then combine the error of the explainer with the approximation error of the model, including spurious interactions created by the model itself.

***Synthetic benchmark for interaction recovery***

The synthetic benchmark uses eight input features that are independent and uniform on $[0,1]$, except in the correlated scenario, in which a Gaussian copula with $\rho = 0.8$ links $x_1$ and $x_3$. The eight scenarios have known structure:

- *additive*: $f = x_1 + 2x_2 + x_3 - x_4$;
- *pair*: $f = x_1 + x_2 + 5x_3x_4$;
- *suppressor*: $f = 2x_1 - 5x_1x_2 + x_3$;
- *two pairs*: $f = 3x_1x_2 - 3x_3x_4 + x_5$;
- *xor*: the exclusive or of $x_1$ and $x_2$, plus $0.5x_3$;
- *three-way*: $f = 10x_1x_2x_3$;
- *conditional*: $f = 4x_1x_2$ if $x_5 > 0.5$ and $f = x_1$ otherwise;
- *correlated*: the *pair* function with correlated inputs.

For each scenario and level, we use 3,000 training rows, 2,000 background rows, 20 test instances and five seeds. Each method produces an $8 \times 8$ matrix of pair scores for every instance. Localized CBFI uses $|G_{4,ij}|$ with $n = 200$ and the marginal sampler, and additionally the conditional sampler in the correlated scenario. k-SII is computed with shapiq at second order with a marginal imputer, exact evaluation with a budget of $2^8$ and 100 background rows (Muschalik et al., 2024a). TreeSHAP interaction values are computed exactly for the tree models (Lundberg et al., 2020), and Friedman's $H^2$ statistic, which is global only, is computed on 150 evaluation points (Friedman & Popescu, 2008).

The methods are compared with several metrics. For each instance, we compute the AUROC and the precision@k for separating true pairs from null pairs, where $k$ is the number of true pairs. The false-positive magnitude is the largest null-pair score divided by the standard deviation of $f$. For scenarios with an analytic local interaction $c(x_i - \mu_i)(x_j - \mu_j)$, we report the correlation with that value and the sign agreement. We also record the runtime per instance and, to measure stability, repeat the explanation ten times with resampled background data on a fixed dataset and model, reporting the coefficient of variation of the true-pair scores and Kendall's $\tau$ of the pair ranking.

***Benchmark for mechanism identification and label robustness***

To assess the robustness of the labels, the instance-level labels computed on fitted RF, XGB and MLP models are compared with the labels computed on the true function ($n = 2{,}000$). This setup covers eight scenarios (additive, pair, AND, OR, max, gate, reverse and xor), three seeds, 40 instances and six features. A rule that takes the majority of the per-sample cases serves as the baseline, and the practical threshold $\tau$ ($0.01$–$0.2 \times \mathrm{sd}(f)$) and the relative threshold $\kappa$ (0–0.2) are varied.

Mechanism identification is tested against a ground truth that is defined semantically, from the generating function and the subgroup of the instance, rather than from the rules of the method. For an AND function, $x_1$ is amplified when both conditions are met and gated when the partner condition is unmet. For an OR function with both features active and for $\max(x_1, x_2)$ with both inputs high, $x_1$ is redundant. For $3x_1 \cdot \mathbf{1}[x_2 > 0.5]$, $x_1$ is amplified when $x_2$ is high and gated when $x_2$ is low; for $2x_1 - 3x_1x_2$ with $x_2 > 0.8$, it is reversed; and for an additive function, it is standalone. We use 15 instances per subgroup, three seeds and four levels.

As baselines, k-SII and TreeSHAP interaction values are mapped to the same five labels with the same thresholds. The simplest variant (*sign only*) uses the first-order value of $x_1$ and the pair value of $(x_1, x_2)$, and therefore cannot separate redundant from gated. The second variant (+ *partner sign*) additionally uses the sign of the partner's first-order value; it requires knowing which feature is the partner, information that localized CBFI does not use. The third variant (+ *rest attribution*) applies the CBFI rules themselves to Shapley quantities and needs no partner information. The first-order value of $x_1$ takes the place of $\overline{A}$, the first-order value plus all pair values involving $x_1$ takes the place of $\overline{B}$, and the summed attribution of all other features, consisting of their first-order and pair values, takes the place of $\overline{C}$. The last rule then uses either the sign agreement between this rest attribution and $\overline{A}$ (rest-sign) or the CBFI threshold $\overline{C} \cdot \mathrm{sign}(\overline{A}) \geq |\overline{D}|/2$ (rest-half). For TreeSHAP interaction values, the corresponding quantities are $\phi_{11}$, $\phi_1$ and $\sum_j \phi_j - \phi_1$.

***Real-world datasets, models and explanation settings***

We use eight real datasets (Table 1): wine quality (red; multiclass; Cortez et al., 2009), medical insurance charges (Lantz, 2013), Adult (Becker & Kohavi, 1996), Bank Marketing (Moro, Cortez & Rita, 2014), Diabetes Health Indicators (balanced; Teboul, 2021), California Housing (Pace & Barry, 1997), Bike Sharing (daily; Fanaee-T & Gama, 2014) and Ames Housing (De Cock, 2011).

Categorical variables are one-hot encoded. In Bank Marketing, the variable `duration` is removed because it is known only after the call, and in Ames Housing, columns with more than 20% missing values are dropped, followed by rows with missing values.

**Table 1. Datasets used in this study.**

| Dataset | Task | Instances | Features (after encoding) | Target | Source |
|---|---|---|---|---|---|
| Wine Quality (red) | classification | 1359 | 11 | 6 classes | UCI |
| Medical Insurance | regression | 1338 | 8 | continuous | Lantz 2013 |
| Adult Census Income | classification | 30,162 | 81 | 2 classes | UCI |
| Bank Marketing (bank.csv) | classification | 4311 | 34 | 2 classes | UCI |
| Diabetes Health Indicators (50/50) | classification | 70,692 | 21 | 2 classes | Kaggle / BRFSS 2015 |
| California Housing | regression | 20,640 | 8 | continuous | scikit-learn |
| Bike Sharing (daily) | regression | 731 | 29 | continuous | UCI |
| Ames Housing | regression | 2218 | 226 | continuous | De Cock 2011 |

Six model families are trained: random forest (Breiman, 2001), XGBoost (Chen & Guestrin, 2016), CatBoost (Prokhorenkova et al., 2018), a multilayer perceptron, an RBF support vector machine (Cortes & Vapnik, 1995) and TabNet (Arık & Pfister, 2021). Every model is a scikit-learn pipeline (StandardScaler followed by the estimator) fitted on an 80% training split only, so scaling parameters are never estimated on test data and all explanations operate on the original feature units. Training sets are capped at 20,000 rows (5,000 for the support vector machine), and TabNet uses early stopping on a 10% validation split.

All explanations use 1,000 training rows as background data. Localized CBFI uses $n = 100$, or $n = 200$ in the case studies. SHAP uses the model-agnostic permutation explainer on the same output as CBFI, that is, the prediction or the probability of the predicted class, with $\max(500, 2p + 1)$ evaluations; exact TreeSHAP is additionally timed for the tree models. LIME uses 5,000 perturbations on the same output. Agreement between explanations is measured on absolute values by Spearman and Kendall rank correlation, top-3 overlap and rank-biased overlap (Webber, Moffat & Zobel, 2010).

To quantify the information that the labels add to an attribution score, we explain RF models on the insurance, wine, California Housing and Bike Sharing data (60 test instances, all features, $n = 200$). For every feature of every instance, we record the CBFI label and the exact TreeSHAP quantities: the SHAP value $\phi_i$, its main part $\phi_{ii}$, its interaction part $\phi_i - \phi_{ii}$, and the attribution of the remaining features $\sum_{j \neq i} \phi_j$. We retain the non-inactive labels with a bootstrap confidence of at least 0.8 (812 of 3,360 feature–instance entries) and ask two questions. First, do entries with nearly the same SHAP value (same sign, difference of at most 10%, same dataset) receive different labels? Second, how well can a classifier (gradient boosting with five-fold cross-validation grouped by instance) reconstruct the label from the TreeSHAP quantities?

The experiments were run on Ubuntu 22.04 with 24 CPU cores, 46 GB of memory and an RTX 3090 GPU, using Python 3.10.12, scikit-learn 1.7.1, shap 0.49.1, shapiq 1.4.1, lime 0.2.0.1, xgboost 3.0.5, catboost 1.2.10 and pytorch-tabnet 4.1.0. The code and exact package versions are available in the repository https://github.com/dkumango/Local_CBFI.

## Results

### *Recovery of known interactions on synthetic functions (Q1)*

When the true function is used as the model, localized CBFI recovers the interaction structure essentially exactly (Table 2). Local precision@k is 0.998, and the false-positive magnitude is 0, so no interaction is attributed in the additive scenario. The per-instance $G_{4,ij}$ correlates with the analytic local interaction at $r = 0.991$, and its sign matches the analytic sign in every case. k-SII performs equally well at this level (precision@k 1.000, $r = 0.982$, sign agreement 0.976), which confirms that both methods estimate the same kind of interventional quantity.

**Table 2. Interaction recovery on synthetic functions (mean over 7 scenarios, 5 seeds, 20 instances).**

| Model | Method | Local AUROC | Local precision@k | Max. null value (sd(f)) | Correlation with analytic value | Sign agreement | Runtime (s per instance) | Max. null value, additive (sd(f)) | Global precision@k (H) |
|---|---|---|---|---|---|---|---|---|---|
| **DGP** | CBFI | 0.990 | 0.998 | 0.000 | 0.991 | 1.000 | 0.063 | 0.000 | 1.000 |
| **DGP** | CBFI-cond | 1.000 | 1.000 | 0.000 | | | 0.132 | | 1.000 |
| **DGP** | kSII | 1.000 | 1.000 | 0.000 | 0.982 | 0.976 | 0.015 | 0.000 | 1.000 |
| **RF** | CBFI | 0.981 | 0.891 | 0.050 | 0.954 | 0.998 | 2.076 | 0.069 | 1.000 |
| **RF** | CBFI-cond | 0.978 | 0.790 | 0.061 | | | 2.243 | | 1.000 |
| **RF** | TreeSHAP-IQ | 0.988 | 0.906 | 0.041 | 0.983 | 0.998 | 0.256 | 0.047 | 1.000 |
| **RF** | kSII | 0.993 | 0.944 | 0.033 | 0.964 | 0.978 | 1.835 | 0.042 | 1.000 |
| **XGB** | CBFI | 0.989 | 0.911 | 0.046 | 0.969 | 0.990 | 0.221 | 0.053 | 0.543 |
| **XGB** | CBFI-cond | 0.991 | 0.840 | 0.047 | | | 0.254 | | 0.543 |
| **XGB** | TreeSHAP-IQ | 0.992 | 0.912 | 0.036 | 0.988 | 0.996 | 0.010 | 0.040 | 0.543 |
| **XGB** | kSII | 0.992 | 0.943 | 0.031 | 0.969 | 0.968 | 0.197 | 0.038 | 0.543 |
| **ANN** | CBFI | 0.955 | 0.856 | 0.044 | 0.968 | 1.000 | 0.151 | 0.023 | 0.152 |
| **ANN** | CBFI-cond | 0.969 | 0.840 | 0.034 | | | 0.222 | | 0.152 |
| **ANN** | kSII | 0.988 | 0.948 | 0.022 | 0.975 | 0.976 | 0.089 | 0.012 | 0.152 |

*Notes:* Model: DGP, the true generating function used as the model. Max. null value: largest |interaction value| assigned to pairs without interaction, in units of the standard deviation of the model output. CBFI-cond: conditional background sampling. Empty cells: not applicable.

On fitted models, localized CBFI recovers the true pairs in most instances (Fig. 2, Table 2), with a local AUROC of 0.96–0.99 and a local precision@k of 0.86 for the MLP, 0.89 for RF and 0.91 for XGB. It is, however, slightly less precise than the dedicated indices: k-SII reaches 0.94–0.95 and TreeSHAP-IQ reaches 0.91 on the tree models, and the largest spurious pair score is somewhat higher for CBFI (0.04–0.05 $\mathrm{sd}(f)$ versus 0.02–0.03 for k-SII). The difference arises mainly in scenarios with several simultaneous interactions (two pairs and conditional), in which the Monte Carlo noise of CBFI ($n = 200$) competes with weak true signals. In the correlated scenario, conditional sampling did not improve CBFI but lowered precision@k (RF 0.79, XGB 0.84), a result we return to in the analysis of sampling strategies.

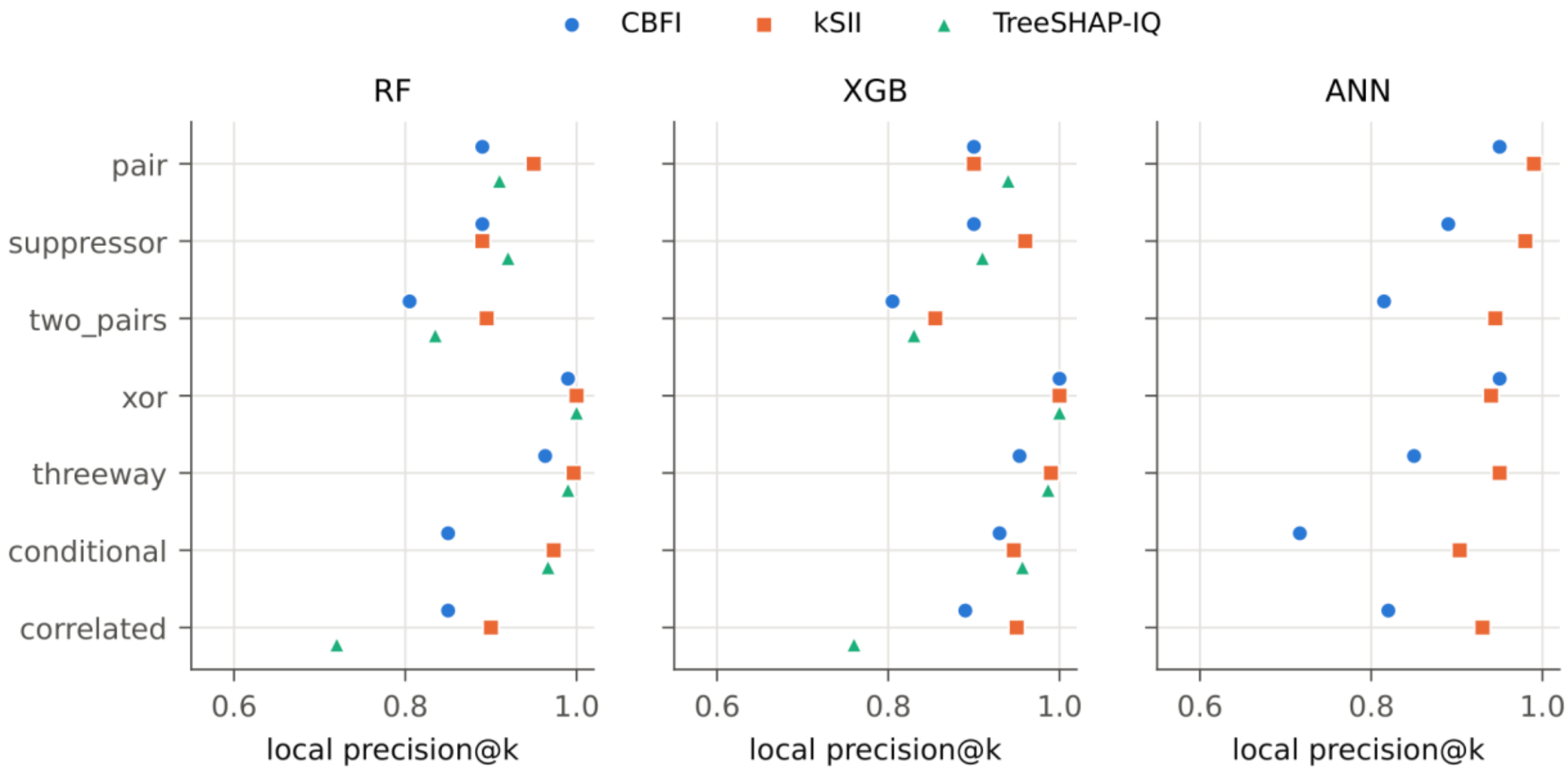

**Figure 2. Recovery of known interacting pairs on synthetic functions.** Each panel shows local precision@k (k = number of true pairs) for one fitted model: RF, XGB or ANN. Values are averaged over 5 seeds and 20 instances. CBFI uses $|G4_{ij}|$ (n = 200), k-SII uses exact shapiq values (budget $2^8$), and TreeSHAP-IQ uses exact TreeSHAP interaction values (tree models only).

When the per-instance scores are averaged, CBFI, k-SII and TreeSHAP-IQ all rank the true pairs first in every scenario and at every level (global precision@k = 1). Friedman's $H^2$ does so for the true function and RF, but its precision drops to 0.54 for XGB and 0.15 for the MLP, and it costs about 110 s per RF model. On the fitted models, computing the full $8 \times 8$ pair matrix takes 0.15–2.1 s per instance with CBFI, which is comparable to model-agnostic k-SII (0.09–1.8 s); exact TreeSHAP-IQ is one to two orders of magnitude faster on tree models (0.01–0.26 s; Fig. 3). Under resampled background data, the true-pair scores of CBFI vary less than those of k-SII (Table 3): the coefficient of variation is 0.27 versus 0.31, and Kendall's $\tau$ of the pair ranking across repeats is 0.86 versus 0.77. TreeSHAP-IQ is deterministic.

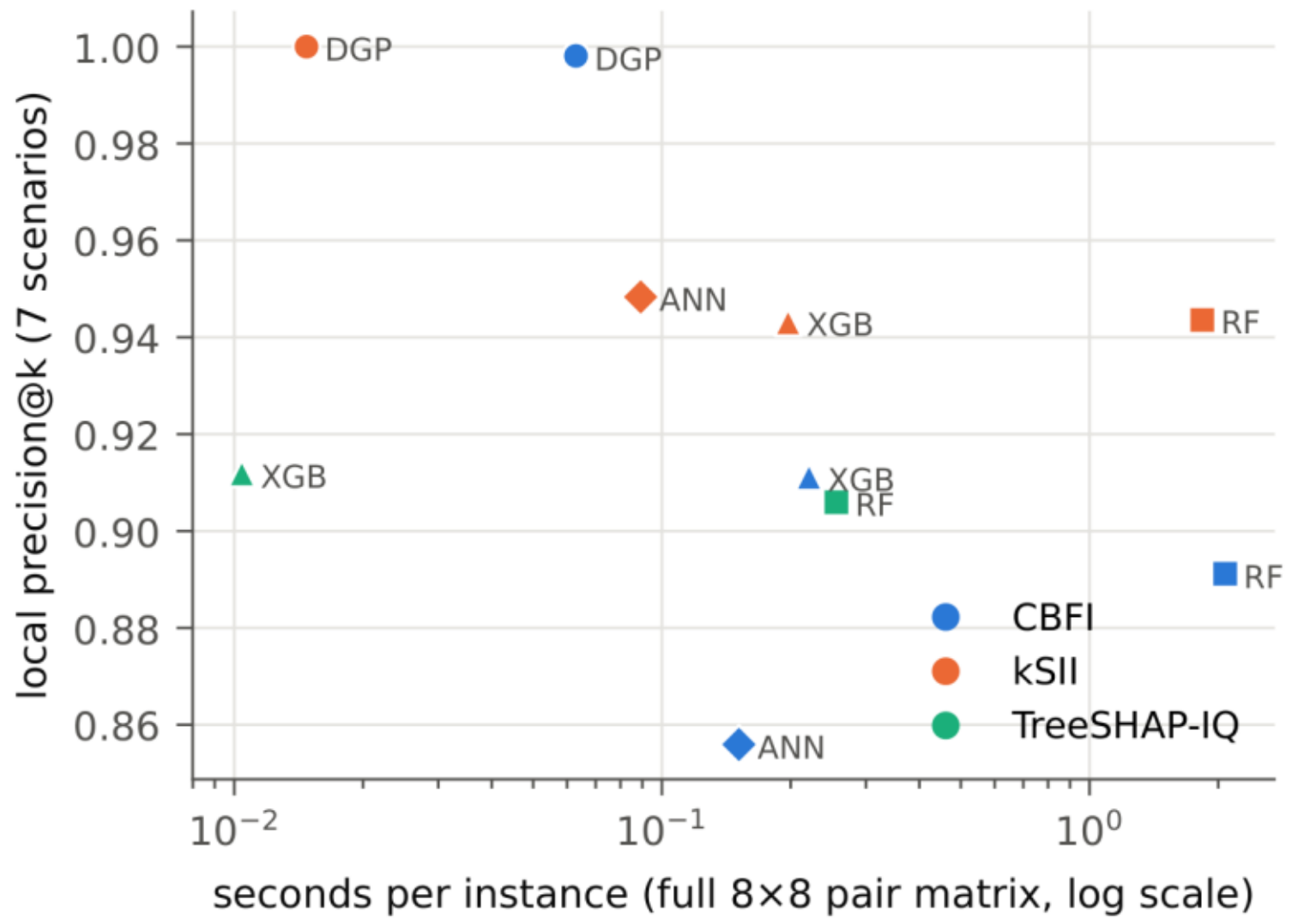

**Figure 3. Accuracy versus runtime.** Local precision@k averaged over the seven interaction scenarios, plotted against seconds per instance for computing the full 8 × 8 pair matrix (log scale). Marker shape indicates the model; DGP denotes the data-generating process, that is, the true function used as the model.

**Table 3. Explainer stability under resampled background data (two_pairs, RF, 10 instances × 10 repeats).**

| Method | CV of true-pair values | Kendall τ of pair ranking |
|---|---|---|
| **CBFI** | 0.272 | 0.856 |
| **kSII** | 0.311 | 0.773 |
| **TreeSHAP-IQ** | 0.000 | 1.000 |

*Notes:* CV, coefficient of variation.

In summary, localized CBFI recovers known interactions exactly on the true function and, on fitted models, with an accuracy close to, but not above, that of dedicated interaction indices at a comparable model-agnostic cost.

### *Identification of interaction mechanisms (Q2)*

Interaction values alone do not reveal the mechanism, because two instances can have identical pairwise interaction values while the underlying logic is opposite. For $f = \mathbf{1}[x_1 > 0.5 \vee x_2 > 0.5]$ at an instance where both features are high, each feature is individually sufficient, which is substitution. For $f = \mathbf{1}[x_1 > 0.5 \wedge x_2 > 0.5]$ at an instance where $x_1$ is high and $x_2$ is low, $x_1$ is switched off by the unmet partner condition, which is gating. Yet the pairwise Shapley interaction is $-0.25$ in both cases: $v(\{1,2\}) - v(\{1\}) - v(\{2\}) + v(\emptyset) = 1 - 1 - 1 + 0.75$ for OR and $0 - 0.5 - 0 + 0.25$ for AND.

The same limitation appears empirically on fitted models (Fig. 4, Table 4). When the sign and magnitude of k-SII or TreeSHAP interaction values are mapped to mechanisms, amplified, reversed and standalone cases are recognized (accuracy 0.99–1.00), but redundant and gated cases are never told apart (accuracy 0), so the overall accuracy is only 0.49–0.50. Localized CBFI labels each feature from its own standalone, in-context and remaining-feature effects, without being told which feature is its partner. It identifies amplified cases with an accuracy of 0.98, reversed cases with 1.00 and standalone cases with 0.97, and it separates gated (0.76) from redundant (0.74) cases, giving an overall accuracy of 0.86–0.88. When the baselines are additionally given the identity of the true partner and the sign of its first-order effect, they become more accurate (0.91–0.94), mainly on redundant cases (0.99).

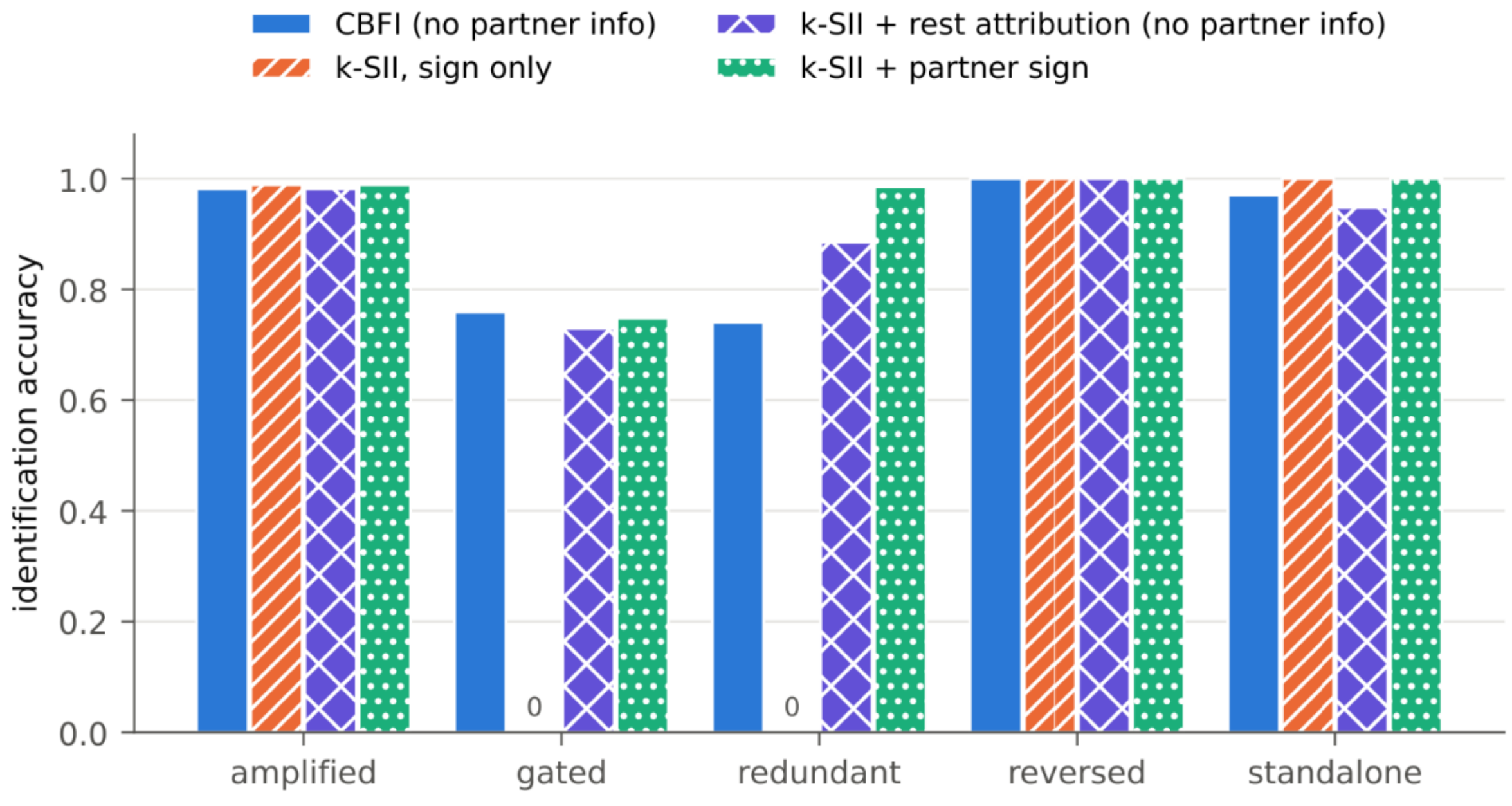

**Figure 4. Identifying how a feature interacts.** Accuracy of identifying the interaction mechanism of $x_1$, with semantic ground truth taken from the generating function and the instance's subgroup. Results are pooled over RF, XGB and ANN and three seeds. CBFI labels each feature on its own, without knowing which feature is its partner. "Sign only" maps first-order and pairwise k-SII values to mechanisms and therefore cannot separate redundant from gated cases (accuracy 0), and "+ partner sign" additionally uses the first-order value of the true partner. "+ rest attribution" applies the CBFI rule to k-SII values, with the summed attribution of all other features in place of the remaining-feature effect, and uses no partner information. The rest-half variant and the TreeSHAP variants are reported in Table 4.

**Table 4. Accuracy of identifying the interaction mechanism of x1 (semantic ground truth; RF, XGB, ANN; 3 seeds). sign: first-order and pair value of x1; rest-sign / rest-half: the CBFI rule applied to Shapley quantities, with the attribution of all other features in place of C (no partner information); +partner: sign of the true partner's first-order value.**

| Method | RF | XGB | ANN | Amplified | Gated | Redundant | Reversed | Standalone |
|---|---|---|---|---|---|---|---|---|
| **CBFI** | 0.856 | 0.881 | 0.864 | 0.981 | 0.759 | 0.741 | 1.000 | 0.970 |
| **kSII (sign)** | 0.500 | 0.500 | 0.492 | 0.989 | 0.000 | 0.000 | 1.000 | 1.000 |
| **kSII (rest-half)** | 0.839 | 0.864 | 0.856 | 0.981 | 0.759 | 0.696 | 1.000 | 0.948 |
| **kSII (rest-sign)** | 0.881 | 0.906 | 0.892 | 0.981 | 0.730 | 0.885 | 1.000 | 0.948 |
| **kSII (+partner)** | 0.936 | 0.942 | 0.914 | 0.989 | 0.748 | 0.985 | 1.000 | 1.000 |
| **TreeSHAP-IQ (sign)** | 0.500 | 0.500 | | 1.000 | 0.000 | 0.000 | 1.000 | 1.000 |
| **TreeSHAP-IQ (rest-half)** | 0.775 | 0.767 | | 1.000 | 0.806 | 0.578 | 0.400 | 1.000 |
| **TreeSHAP-IQ (rest-sign)** | 0.800 | 0.806 | | 1.000 | 0.783 | 0.728 | 0.400 | 1.000 |
| **TreeSHAP-IQ (+partner)** | 0.936 | 0.942 | | 1.000 | 0.761 | 0.994 | 1.000 | 1.000 |

*Notes:* Columns RF–ANN: accuracy per model; columns Amplified–Standalone: accuracy per true mechanism.

The decisive ingredient is the effect of the remaining features. When the CBFI rules are applied to k-SII values, with the summed attribution of all other features in place of $\overline{C}$, no partner information is needed, and the accuracy is 0.88–0.91 (rest-sign) or 0.84–0.86 (rest-half), on par with localized CBFI. The same construction on TreeSHAP interaction values reaches 0.77–0.81 and misses most reversed cases (0.40). The mechanism is therefore not hidden from Shapley-type attributions, but it is not expressed by an interaction value. It becomes identifiable through the taxonomy, which compares the standalone and in-context effects of a feature and checks whether the remaining features reproduce the lost effect. Localized CBFI computes the three required quantities directly from four hybrid states, with bootstrap confidence, and the same taxonomy can be layered on Shapley interaction indices.

The labels are also robust on fitted models (Fig. 5, Table 5). Labels obtained by assigning every background sample to a case and taking the majority agree poorly with the labels on the true function (accuracy 0.42–0.52), because the stepwise structure of fitted models makes individual samples look interactive. Averaging the per-sample effects, bootstrapping the means and requiring a practically relevant effect raises the agreement to 0.95 for all three models ($\tau = 0.1\mathrm{sd}(f)$; macro-F1 0.85–0.86). For RF, the recall per label is 0.97 for inactive, 0.91 for standalone, 0.93 for amplified, 0.94 for redundant, 0.96 for gated and 0.80 for reversed. The relative threshold $\kappa$ has almost no influence (accuracy 0.947–0.955 for $\kappa$ = 0–0.2), whereas the practical threshold $\tau$ matters: the accuracy is 0.60–0.70 at $\tau = 0.01$ and 0.97 at $\tau = 0.2$. Because $\tau$ expresses the smallest effect a user considers relevant, we report it with every explanation. The bootstrap confidence flags unreliable labels, with a mean of 0.98 for correct labels and 0.78 for incorrect ones.

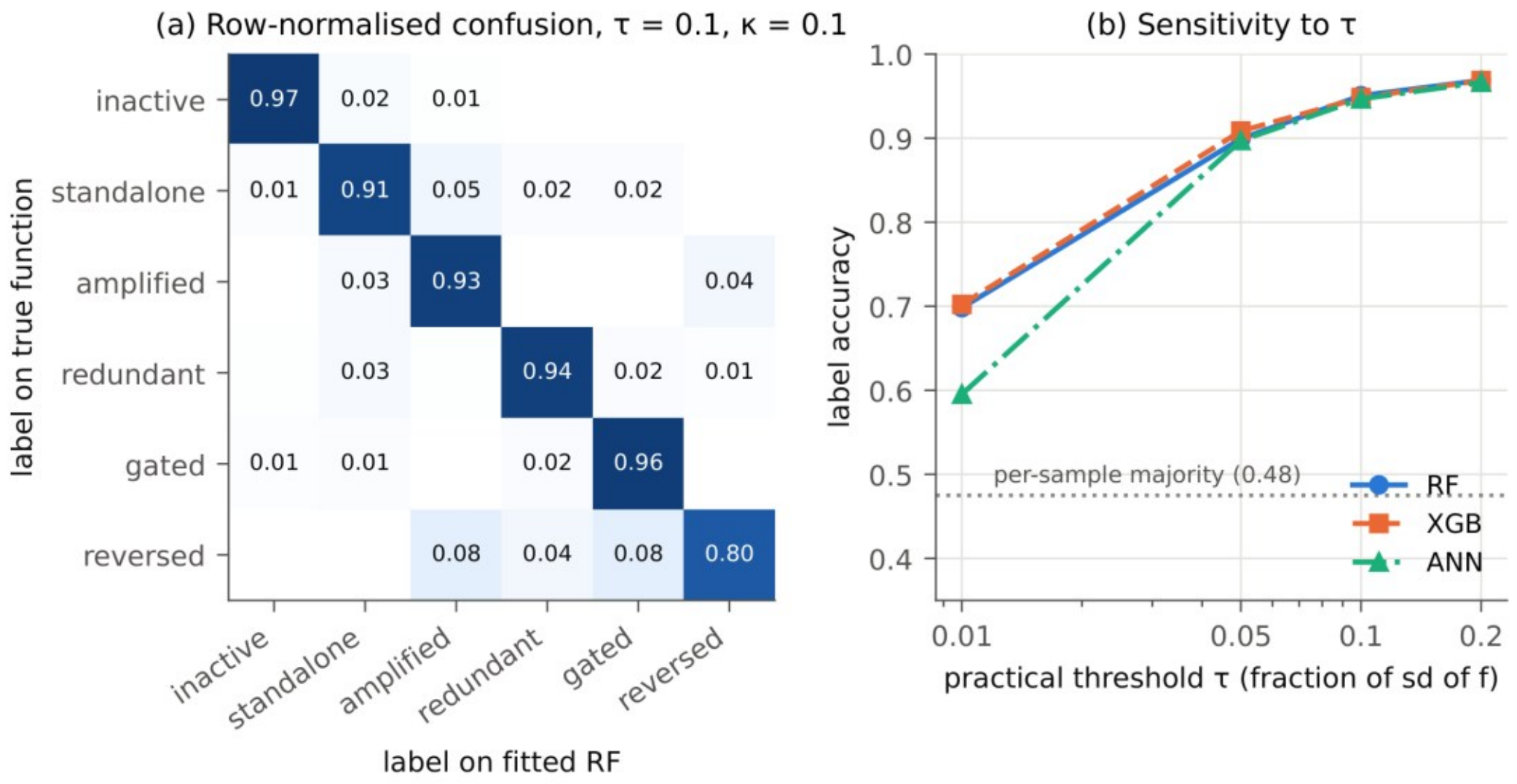

**Figure 5. Robustness of the case labels on fitted models.** (A) Row-normalised confusion between labels on a fitted RF and labels on the true function (τ = 0.1, κ = 0.1; 8 scenarios × 3 seeds × 40 instances × 6 features). (B) Label accuracy as a function of the practical threshold τ, expressed as a fraction of the standard deviation of f. The dotted line shows the per-sample majority rule.

**Table 5. Agreement of case labels on fitted models with labels on the true function (8 scenarios × 3 seeds × 40 instances × 6 features).**

| Labelling rule | Accuracy (RF) | Macro-F1 (RF) | Accuracy (XGB) | Macro-F1 (XGB) | Accuracy (ANN) | Macro-F1 (ANN) |
|---|---|---|---|---|---|---|
| per-sample majority | 0.521 | 0.362 | 0.481 | 0.344 | 0.423 | 0.379 |
| instance tau=0.01 k=0.1 | 0.698 | 0.645 | 0.702 | 0.663 | 0.595 | 0.612 |
| instance tau=0.05 k=0.1 | 0.899 | 0.806 | 0.909 | 0.823 | 0.897 | 0.820 |
| instance tau=0.1 k=0.1 | 0.951 | 0.860 | 0.948 | 0.850 | 0.947 | 0.858 |
| instance tau=0.2 k=0.1 | 0.969 | 0.931 | 0.969 | 0.911 | 0.966 | 0.901 |
| instance tau=0.1 k=0.0 | 0.950 | 0.860 | 0.948 | 0.850 | 0.947 | 0.858 |
| instance tau=0.1 k=0.05 | 0.950 | 0.860 | 0.948 | 0.850 | 0.947 | 0.858 |
| instance tau=0.1 k=0.2 | 0.955 | 0.868 | 0.951 | 0.854 | 0.949 | 0.863 |

*Notes:* τ (tau): practical threshold as a fraction of sd(f); κ (k): relative threshold.

One caveat concerns correlated duplicates. When $x_2$ is a near-copy of $x_1$, fitted models rely mainly on one copy, and XGB labels $x_2$ inactive in 72% of instances. The labels then describe the actual reliance of the model rather than the correlation in the data, so correlated features are not automatically labelled redundant. Overall, for every instance and feature, localized CBFI states whether the effect is standalone or context-dependent and, if it is context-dependent, whether the context amplifies, substitutes for, switches off or reverses the feature. On fitted models, these labels agree with the generating logic in 95% of cases at the default threshold.

### *Stability, computational cost and scalability (Q3)*

Across six datasets, six models and five instances with 30 repeats each, the standard deviation of $G_4$ relative to |Imp| decreases at roughly the $1/\sqrt{n}$ rate expected for Monte Carlo estimation (Fig. 6A, Table 6): it is 0.25 at $n = 10$, 0.08 at $n = 100$ and 0.03 at $n = 500$. At $n = 100$, the median deviation from a reference run with $n = 1{,}000$ is 1.9% of |Imp|, and the variability is similar across

models (0.07–0.10). We therefore use $n = 100$ for the multi-domain experiments and $n = 200$ for the case studies. The variability is normalized by |Imp| rather than by $G_4$ itself, because $G_4$ can legitimately be close to zero, which would make its own coefficient of variation unstable.

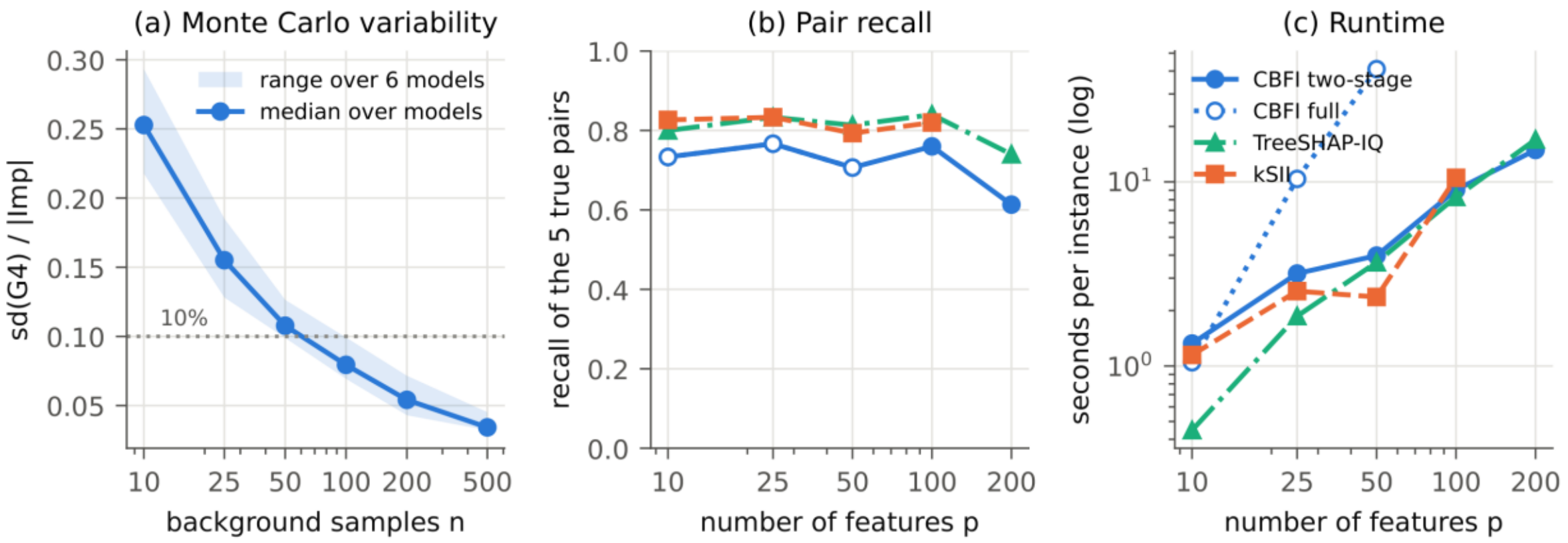

**Figure 6. Stability and scalability.** (A) Monte Carlo variability of G4 relative to |Imp| as a function of the number of background samples n. The line is the median over six models, the band is the range over models, and the data cover 6 datasets × 5 instances × 30 repeats. (B) Recall of the five true pairs as the number of features p grows. (C) Runtime per instance (log scale). The CBFI two-stage procedure screens features by mean |d| and then computes $G4_{ij}$ only among the top 12; CBFI full computes all pairs and is run only up to p = 50.

**Table 6. Monte Carlo variability of localized CBFI vs number of background samples (6 datasets × 6 models × 5 instances × 30 repeats; medians).**

| Background samples n | CV of G4 (relative to \|Imp\|) | CV of Imp | Median \|deviation\| from n = 1,000 (relative to \|Imp\|) |
|---|---|---|---|
| **10.000** | 0.252 | 0.268 | 0.035 |
| **25.000** | 0.151 | 0.169 | 0.023 |
| **50.000** | 0.108 | 0.117 | 0.022 |
| **100.0** | 0.082 | 0.084 | 0.019 |
| **200.0** | 0.054 | 0.057 | 0.019 |
| **500.0** | 0.034 | 0.034 | 0.019 |

*Notes:* CV, coefficient of variation.

Explaining all features of an instance requires about $3np$ model evaluations (Table 7). Over six datasets, this took 1.1–2.8 s per instance for XGBoost, CatBoost, the MLP, the support vector machine and TabNet, and 10.4 s for the random forest (32 s on Ames Housing with 226 features). Localized CBFI is thus 1.4 (support vector machine) to 15 (random forest) times slower than permutation SHAP with its default evaluation budget, and much slower than exact TreeSHAP on tree models (0.01–0.55 s).

**Table 7. Mean runtime per explained instance (s), averaged over six datasets.**

| Model | Localized CBFI (s) | Permutation SHAP (s) | LIME (s) | TreeSHAP (s) |
|---|---|---|---|---|
| **ANN** | 1.100 | 0.213 | 0.138 | |
| **CAT** | 1.145 | 0.212 | 0.131 | 0.010 |
| **RF** | 10.358 | 0.706 | 0.226 | 0.548 |
| **SVM** | 2.795 | 1.942 | 0.524 | |
| **TabNet** | 2.441 | 0.565 | 0.267 | |
| **XGB** | 1.244 | 0.292 | 0.160 | 0.008 |

*Notes:* Empty cells: TreeSHAP is not available for the model.

To examine scalability, we embedded five interacting pairs among $p$ = 10–200 features (Fig. 6B–C, Table 8). The single-feature screen, which ranks features by their mean $|d_s|$, placed all ten interacting features within its top 12 for every $p$ (screening recall 1.00). Computing $G_{4,ij}$ only among the screened features gave the same pair recall as the full pairwise computation (0.71–0.77 for $p \leq 50$), while its runtime grew roughly linearly (4.0 s versus 41 s at $p = 50$, and 15 s at $p = 200$). The pair recall was 0.06–0.13 lower than that of TreeSHAP-IQ and k-SII, which is consistent with the results on synthetic functions. For high-dimensional data, we therefore recommend the two-stage procedure, in which the features are screened first and the pairs are then analysed among the candidate features.

**Table 8. Scalability: pair recall and runtime vs number of features (5 true pairs; RF and XGB; 3 seeds × 5 instances).**

| Features p | Recall@5: CBFI full | Recall@5: CBFI two-stage | Recall@5: TreeSHAP-IQ | Recall@5: k-SII | Runtime: CBFI full (s) | Runtime: CBFI two-stage (s) | Runtime: TreeSHAP-IQ (s) | Runtime: k-SII (s) | Screening recall (top 12) |
|---|---|---|---|---|---|---|---|---|---|
| **10.000** | 0.733 | 0.733 | 0.800 | 0.827 | 1.049 | 1.325 | 0.448 | 1.149 | 1.000 |
| **25.000** | 0.767 | 0.767 | 0.833 | 0.833 | 10.381 | 3.180 | 1.867 | 2.552 | 1.000 |
| **50.000** | 0.707 | 0.707 | 0.813 | 0.793 | 40.968 | 3.978 | 3.657 | 2.372 | 1.000 |
| **100.0** | | 0.760 | 0.840 | 0.820 | | 9.046 | 8.279 | 10.523 | 1.000 |
| **200.0** | | 0.613 | 0.740 | | | 14.905 | 16.965 | | 1.000 |

*Notes:* Runtime in seconds per explained instance.

## *Sensitivity to the background sampling strategy (Q4)*

Replacing marginal sampling with conditional sampling from the $k$ nearest neighbours ($k$ = 20, 50 and 200, on the two features most correlated with the target feature) changed the feature rankings moderately (Table 9). Kendall's $\tau$ with the marginal ranking ranged from 0.47–0.58 for California Housing to 0.80–0.82 for insurance, and the overlap of the top-3 features was 0.50–0.93. The share of interaction in the explanation changed little, by at most 0.07. The largest changes occurred for strongly correlated features, such as latitude and longitude or the acidity variables in the wine data. In the synthetic correlated scenario, marginal sampling recovered the true pair more accurately. Marginal and conditional sampling answer different questions: marginal sampling describes how the model responds when a feature is changed, whereas conditional sampling describes effects within the data distribution. We therefore keep marginal sampling as the default and recommend reporting both when features are strongly correlated.

**Table 9. Sensitivity to the sampling strategy: agreement of conditional (KNN) with marginal sampling (RF, 20 instances).**

| Dataset | Sampler | Kendall τ of Imp vs marginal | Top-3 overlap of Imp | Interaction share |
|---|---|---|---|---|
| **adult** | conditional-k20 | 0.545 | 0.650 | 0.662 |
| **adult** | conditional-k200 | 0.650 | 0.717 | 0.679 |
| **adult** | conditional-k50 | 0.585 | 0.650 | 0.662 |
| **bike** | conditional-k20 | 0.614 | 0.517 | 0.437 |
| **bike** | conditional-k200 | 0.702 | 0.667 | 0.471 |
| **bike** | conditional-k50 | 0.633 | 0.500 | 0.440 |
| **california** | conditional-k20 | 0.468 | 0.617 | 0.589 |
| **california** | conditional-k200 | 0.582 | 0.633 | 0.649 |
| **california** | conditional-k50 | 0.500 | 0.633 | 0.634 |
| **diabetes** | conditional-k20 | 0.683 | 0.700 | 0.689 |
| **diabetes** | conditional-k200 | 0.761 | 0.733 | 0.767 |
| **diabetes** | conditional-k50 | 0.724 | 0.750 | 0.756 |
| **insurance** | conditional-k20 | 0.812 | 0.933 | 0.472 |
| **insurance** | conditional-k200 | 0.817 | 0.917 | 0.505 |
| **insurance** | conditional-k50 | 0.801 | 0.900 | 0.482 |
| **wine** | conditional-k20 | 0.533 | 0.767 | 0.703 |
| **wine** | conditional-k200 | 0.595 | 0.800 | 0.758 |
| **wine** | conditional-k50 | 0.560 | 0.767 | 0.732 |

### *Case studies and multi-domain analysis on real data (Q5)*

**Medical insurance.** For the RF model (test $R^2 = 0.864$), smoking status has the largest in-context importance (mean |Imp| = 8,207 USD), and most of it is standalone (mean $|G_1|$ = 7,776 USD; Fig. 7A). Its interaction component changes sign with BMI (Fig. 8A). For smokers with a BMI of 30 or more, the context amplifies the effect of smoking (mean $G_4$ = +7,198 USD over seven instances; in the label-robustness run on 30 test instances, all five obese smokers were labelled amplified), whereas for smokers with a BMI below 30 it attenuates the effect (mean $G_4$ = −8,804 USD). The pairwise analysis identifies BMI–smoker as by far the strongest interaction (mean $|G_{4,ij}|$ = 3,464 USD), followed by age–children (899 USD) and age–BMI (788 USD) (Fig. 9A). As an independent check, the in-context importance of smoking agrees with the counterfactual effect of flipping smoking status while holding all other features fixed (Pearson $r = 0.963$; Fig. 8B). These statements describe the fitted model and are not causal claims about medical costs. The structure they reveal, a large standalone effect of smoking whose size depends on obesity, is nevertheless consistent with the obesity–smoking interaction commonly modelled for this dataset (Lantz, 2013).

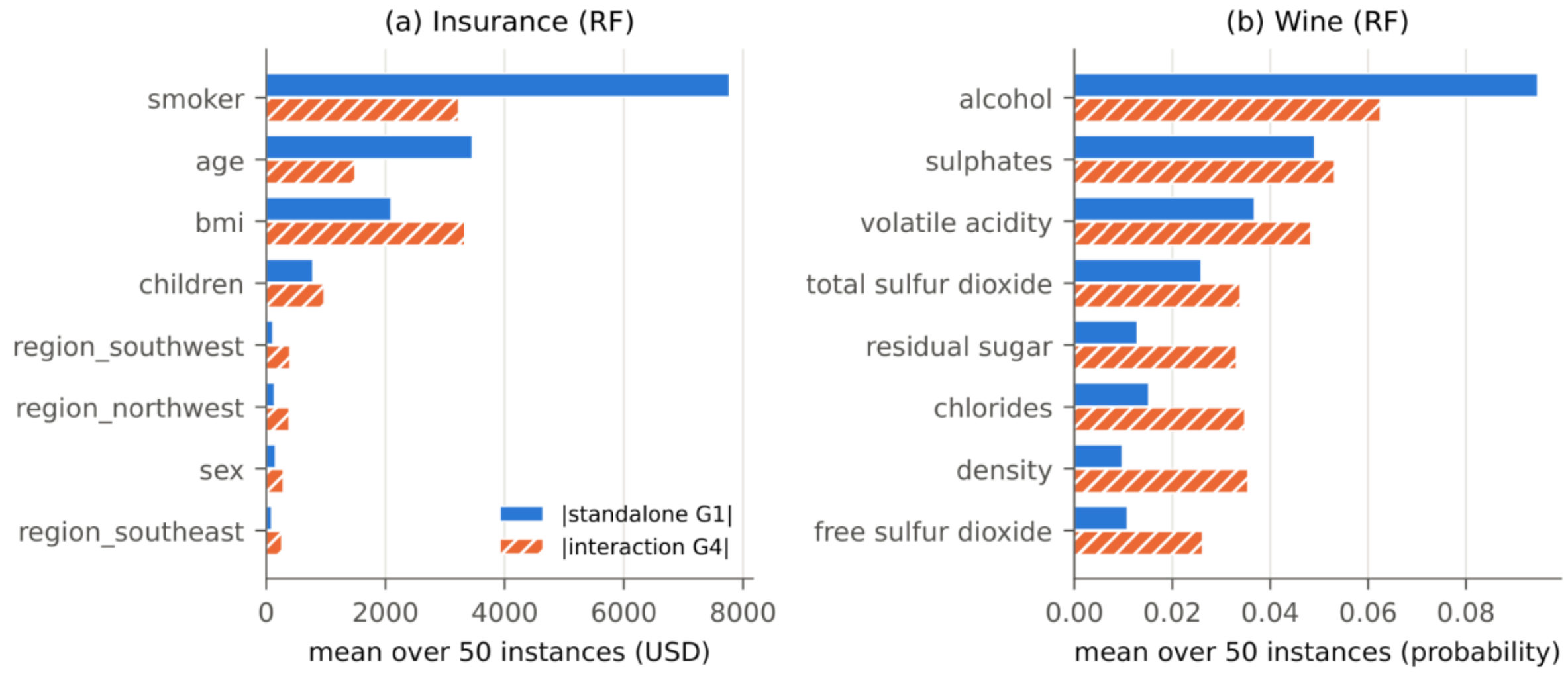

**Figure 7. Standalone and interaction components in the case studies.** Mean |G1| and mean |G4| over 50 test instances for (A) medical insurance (USD) and (B) wine quality (probability of the predicted class). Model: RF.

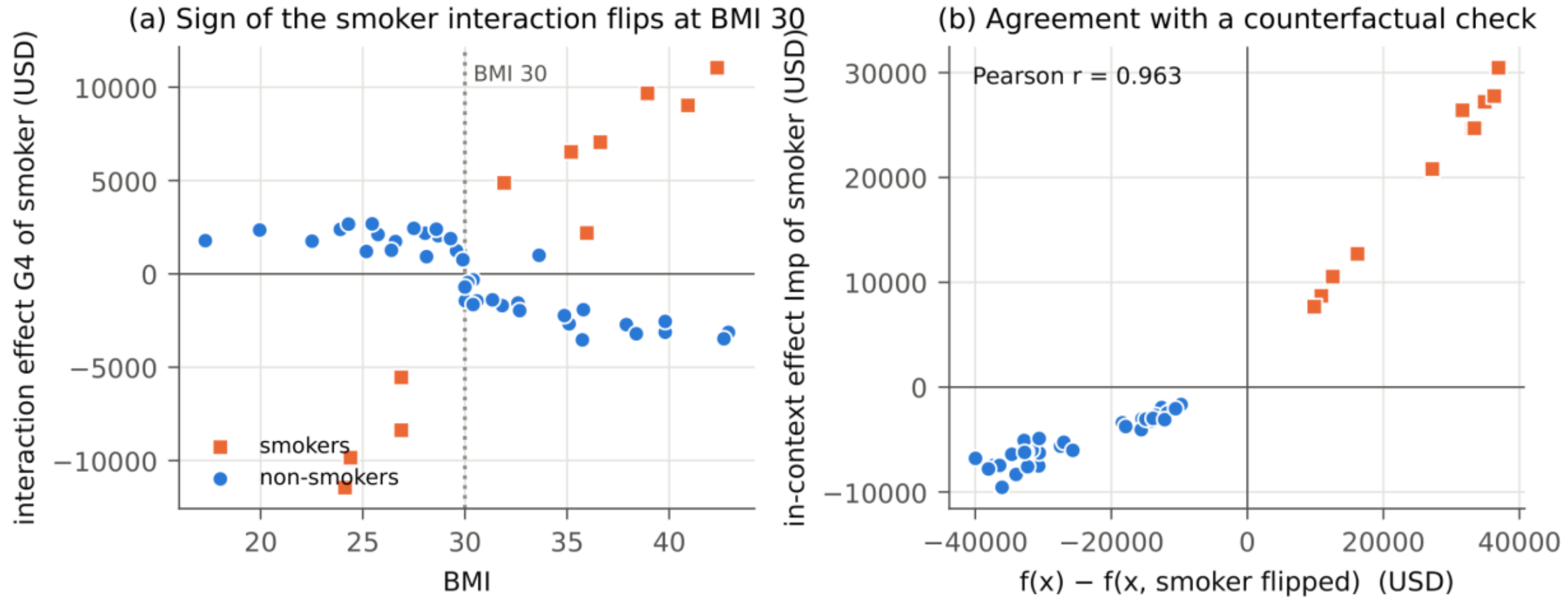

**Figure 8. Medical insurance: smoker × BMI.** (A) Interaction effect G4 of smoker for 50 test instances, plotted against BMI. The sign of the interaction flips at BMI = 30. (B) The in-context effect Imp of smoker agrees with a counterfactual check, f(x) − f(x with smoker flipped) (Pearson r = 0.963). The two quantities are not expected to coincide, because Imp compares against the background distribution of smoking status rather than a full flip.

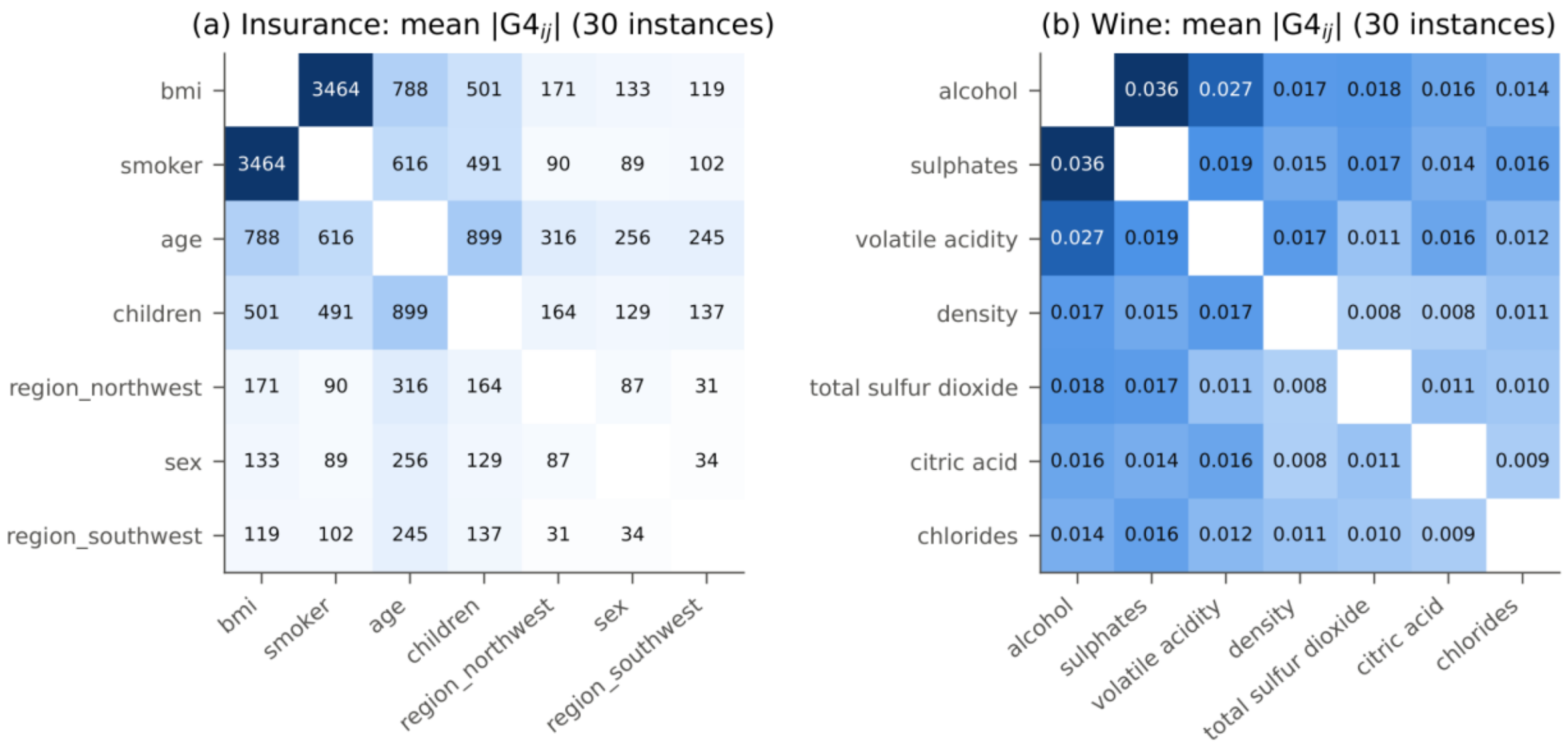

**Figure 9. Pairwise interaction strength.** Mean $|G4_{ij}|$ over 30 test instances for (A) medical insurance and (B) wine quality. The matrix is symmetric (undirected), and the seven features with the largest totals are shown.

**Wine quality.** For the multiclass RF (accuracy 0.618), alcohol has the largest importance and acts mainly through its standalone effect (mean $G_1$ = +0.074; labelled standalone in 43% of instances; Fig. 7B). Varying alcohol alone changes the probability of the predicted class by 0.29 on average, as measured by individual conditional expectation, so alcohol does not act only in combination with other features. The strongest pair is alcohol–sulphates (mean $|G_{4,ij}|$ = 0.036; Fig. 9B). At the default threshold, most minor features are labelled inactive in 60–80% of instances.

**Agreement with SHAP and LIME.** On the same output, the rankings by |Imp| agree moderately with SHAP (Kendall $\tau$ 0.53 for wine and 0.67 for insurance) and less with LIME (0.30 and 0.55), while SHAP and LIME agree with each other to a similar degree (0.50 and 0.70) (Table 10). The standalone component $G_1$ agrees with SHAP more than the interaction component $G_4$ does (0.47 versus 0.28 for wine and 0.75 versus 0.53 for insurance). This is expected, since an additive attribution has no separate place for the interaction part. We report these agreements only descriptively, for the reasons given in the Discussion.

**Table 10. Rank agreement between explanations (|values|; mean over 50 instances; RF).**

| Comparison | Dataset | Spearman ρ | Kendall τ | Top-3 Jaccard | RBO |
|---|---|---|---|---|---|
| **G1 vs SHAP** | wine | 0.595 | 0.470 | 0.564 | 0.832 |
| **G4 vs SHAP** | wine | 0.371 | 0.283 | 0.348 | 0.738 |
| **Imp vs LIME** | wine | 0.388 | 0.295 | 0.452 | 0.763 |
| **Imp vs SHAP** | wine | 0.649 | 0.529 | 0.624 | 0.843 |
| **SHAP vs LIME** | wine | 0.626 | 0.495 | 0.622 | 0.835 |
| **G1 vs SHAP** | insurance | 0.862 | 0.746 | 0.830 | 0.941 |
| **G4 vs SHAP** | insurance | 0.669 | 0.534 | 0.632 | 0.839 |
| **Imp vs LIME** | insurance | 0.689 | 0.550 | 0.680 | 0.884 |
| **Imp vs SHAP** | insurance | 0.782 | 0.671 | 0.770 | 0.915 |
| **SHAP vs LIME** | insurance | 0.829 | 0.701 | 0.800 | 0.929 |

*Notes:* RBO, rank-biased overlap.

**Deletion fidelity.** Replacing the top-ranked features with background values degrades the prediction much faster than a random ranking for all three methods, and the methods differ only slightly (Fig. 10, Table 11). The mean output change over $k$ is 0.155 for SHAP, 0.148 for LIME, 0.146 for CBFI and 0.097 for the random ranking on the wine data, and 8,113 USD for CBFI, 8,027 USD for SHAP, 7,971 USD for LIME and 5,075 USD for the random ranking on the insurance data. In this deletion test, the importance ranking of localized CBFI is thus comparably faithful to the model as the established attributions; its additional value lies in the decomposition and the labels rather than in a better ranking.

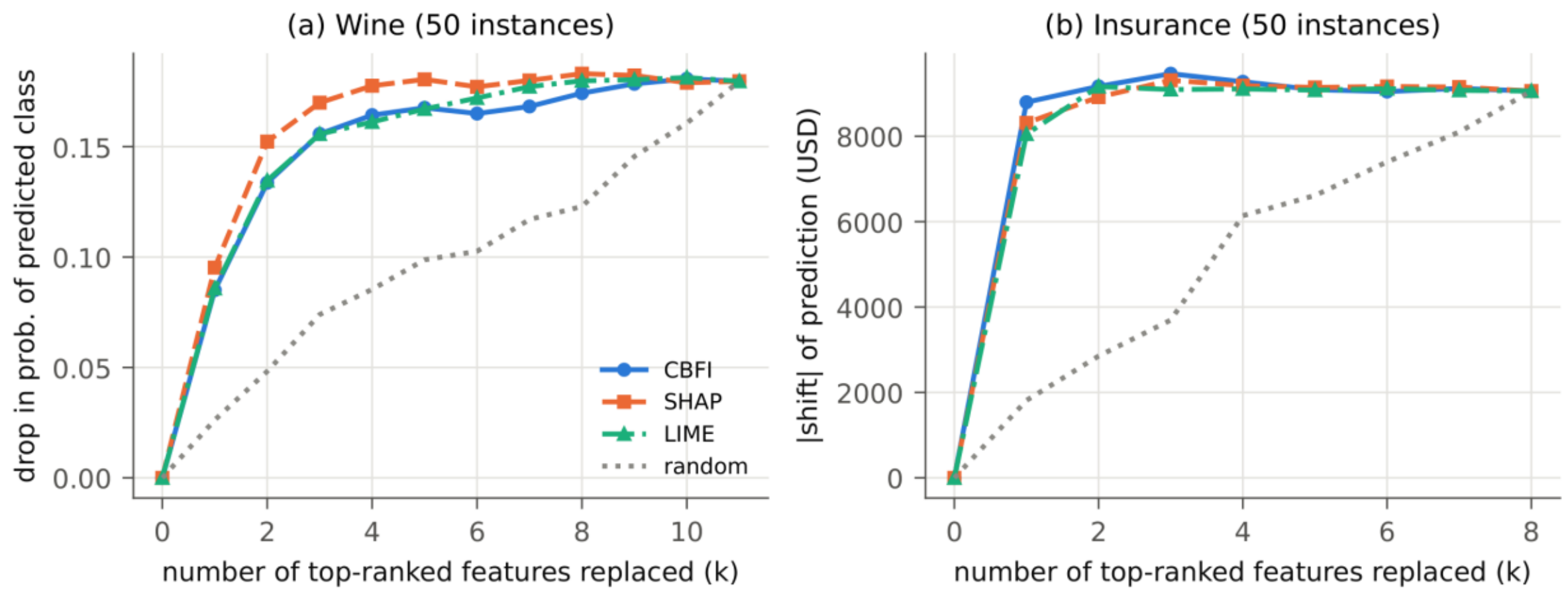

**Figure 10. Deletion fidelity.** The top-k features ranked by each explanation are replaced with background values, and the change in model output is averaged over 50 instances: (A) drop in the probability of the predicted class for wine; (B) absolute shift of the prediction for insurance. A random ranking is shown as a reference.

**Table 11. Deletion fidelity: mean output change over k (higher = the ranking removes more of the decision; 50 instances).**

| Dataset (output unit) | CBFI G1 | CBFI Imp | LIME | SHAP | Random |
|---|---|---|---|---|---|
| **wine** | 0.145 | 0.146 | 0.148 | 0.155 | 0.097 |
| **insurance** | 7,916 | 8,113 | 7,971 | 8,027 | 5,075 |

*Notes:* Output unit: probability of the predicted class (wine); USD (insurance).

**Multi-domain results.** Across six datasets and six model types (Table 12), the interaction component accounts for 42–79% of the explained effect, measured as the mean $|d_s|$ relative to the sum of mean $|G_1|$ and mean $|d_s|$. The interaction activation rate differs clearly between domains. Interaction is active for most features and instances in California Housing (0.78–0.90), where the location features act jointly, but rarely in the one-hot-encoded Adult and Ames Housing data (0.07–0.17). Within each dataset, the activation rate is similar across model types; on California Housing, for example, it is 0.78–0.90 while $R^2$ ranges from 0.69 to 0.85. The interaction share varies more (for example, 0.52–0.79 on Bank Marketing), so it reflects the model as well as the domain. The decomposition thus provides a domain-level diagnostic of whether the predictions of a model are dominated by standalone or by context-dependent effects.

**Table 12. Multi-domain results (10 test instances per dataset-model pair; n = 100).**

| Dataset | Model | p | Training rows | Test score | Interaction share | Activation rate | CBFI (s) | Perm. SHAP (s) | LIME (s) | TreeSHAP (s) | Imp–SHAP Kendall τ | Imp–SHAP RBO |
|---|---|---|---|---|---|---|---|---|---|---|---|---|
| **adult** | RF | 81 | 20,000 | 0.849 | 0.687 | 0.107 | 13.833 | 0.776 | 0.376 | 1.384 | 0.692 | 0.688 |
| **adult** | XGB | 81 | 20,000 | 0.864 | 0.592 | 0.097 | 1.128 | 0.580 | 0.329 | 0.004 | 0.749 | 0.774 |
| **adult** | CAT | 81 | 20,000 | 0.864 | 0.588 | 0.101 | 1.079 | 0.522 | 0.313 | 0.012 | 0.753 | 0.822 |
| **adult** | ANN | 81 | 20,000 | 0.847 | 0.584 | 0.102 | 0.960 | 0.560 | 0.329 | | 0.673 | 0.764 |
| **adult** | SVM | 81 | 5000 | 0.841 | 0.585 | 0.108 | 3.521 | 1.638 | 0.808 | | 0.661 | 0.747 |
| **adult** | TabNet | 81 | 20,000 | 0.845 | 0.606 | 0.101 | 2.739 | 0.551 | 0.324 | | 0.605 | 0.698 |
| **bank** | RF | 34 | 3448 | 0.888 | 0.759 | 0.298 | 6.187 | 0.926 | 0.156 | 0.112 | 0.541 | 0.638 |
| **bank** | XGB | 34 | 3448 | 0.891 | 0.696 | 0.278 | 0.321 | 0.606 | 0.086 | 0.005 | 0.620 | 0.711 |
| **bank** | CAT | 34 | 3448 | 0.893 | 0.633 | 0.283 | 0.288 | 0.541 | 0.057 | 0.009 | 0.640 | 0.781 |
| **bank** | ANN | 34 | 3448 | 0.884 | 0.519 | 0.287 | 0.258 | 0.542 | 0.062 | | 0.587 | 0.803 |
| **bank** | SVM | 34 | 3448 | 0.878 | 0.706 | 0.262 | 0.783 | 1.314 | 0.310 | | 0.422 | 0.622 |
| **bank** | TabNet | 34 | 3448 | 0.884 | 0.788 | 0.295 | 1.050 | 0.280 | 0.154 | | 0.442 | 0.682 |
| **diabetes** | RF | 21 | 20,000 | 0.738 | 0.737 | 0.463 | 3.921 | 0.591 | 0.240 | 0.996 | 0.540 | 0.723 |
| **diabetes** | XGB | 21 | 20,000 | 0.745 | 0.634 | 0.454 | 0.185 | 0.157 | 0.120 | 0.007 | 0.531 | 0.798 |
| **diabetes** | CAT | 21 | 20,000 | 0.747 | 0.587 | 0.451 | 0.156 | 0.077 | 0.091 | 0.011 | 0.587 | 0.784 |
| **diabetes** | ANN | 21 | 20,000 | 0.746 | 0.571 | 0.449 | 0.130 | 0.022 | 0.091 | | 0.550 | 0.797 |
| **diabetes** | SVM | 21 | 5000 | 0.740 | 0.590 | 0.454 | 0.880 | 2.636 | 0.655 | | 0.632 | 0.779 |
| **diabetes** | TabNet | 21 | 20,000 | 0.743 | 0.575 | 0.428 | 0.642 | 0.285 | 0.150 | | 0.531 | 0.817 |
| **california** | RF | 8 | 16,512 | 0.807 | 0.650 | 0.883 | 1.268 | 1.482 | 0.169 | 0.650 | 0.721 | 0.912 |
| **california** | XGB | 8 | 16,512 | 0.841 | 0.609 | 0.894 | 0.048 | 0.233 | 0.057 | 0.007 | 0.607 | 0.895 |
| **california** | CAT | 8 | 16,512 | 0.845 | 0.604 | 0.873 | 0.047 | 0.063 | 0.036 | 0.009 | 0.664 | 0.918 |
| **california** | ANN | 8 | 16,512 | 0.790 | 0.596 | 0.904 | 0.041 | 0.083 | 0.037 | | 0.650 | 0.887 |
| **california** | SVM | 8 | 5000 | 0.750 | 0.517 | 0.881 | 0.276 | 4.818 | 0.508 | | 0.843 | 0.955 |
| **california** | TabNet | 8 | 16,512 | 0.688 | 0.536 | 0.780 | 0.208 | 0.630 | 0.103 | | 0.686 | 0.904 |
| **bike** | RF | 29 | 584 | 0.872 | 0.457 | 0.282 | 4.436 | 0.345 | 0.097 | 0.029 | 0.707 | 0.891 |
| **bike** | XGB | 29 | 584 | 0.875 | 0.503 | 0.294 | 0.237 | 0.087 | 0.068 | 0.006 | 0.723 | 0.867 |
| **bike** | CAT | 29 | 584 | 0.891 | 0.428 | 0.299 | 0.226 | 0.021 | 0.039 | 0.009 | 0.730 | 0.893 |
| **bike** | ANN | 29 | 584 | 0.856 | 0.501 | 0.316 | 0.192 | 0.017 | 0.039 | | 0.610 | 0.823 |
| **bike** | SVM | 29 | 584 | 0.882 | 0.523 | 0.311 | 0.359 | 0.338 | 0.136 | | 0.596 | 0.809 |
| **bike** | TabNet | 29 | 584 | 0.857 | 0.537 | 0.305 | 1.251 | 1.346 | 0.457 | | 0.623 | 0.832 |
| **ames** | RF | 226 | 1774 | 0.906 | 0.550 | 0.073 | 32.501 | 0.118 | 0.317 | 0.120 | 0.669 | 0.760 |
| **ames** | XGB | 226 | 1774 | 0.912 | 0.548 | 0.095 | 5.543 | 0.086 | 0.302 | 0.016 | 0.682 | 0.753 |
| **ames** | CAT | 226 | 1774 | 0.936 | 0.418 | 0.088 | 5.077 | 0.049 | 0.252 | 0.008 | 0.804 | 0.815 |
| **ames** | ANN | 226 | 1774 | 0.816 | 0.660 | 0.174 | 5.019 | 0.054 | 0.271 | | 0.561 | 0.541 |
| **ames** | SVM | 226 | 1774 | 0.784 | 0.477 | 0.134 | 10.955 | 0.906 | 0.725 | | 0.593 | 0.810 |
| **ames** | TabNet | 226 | 1774 | 0.847 | 0.582 | 0.136 | 8.754 | 0.296 | 0.414 | | 0.524 | 0.594 |

*Notes:* Test score: accuracy (classification) or $R^2$ (regression). Runtime in seconds per explained instance. Empty cells: TreeSHAP is not available for the model.

**Information beyond the attribution score.** We finally asked whether the label tells a user something that the attribution score does not (Fig. 11, Table 13). Among pairs of feature–instance entries with nearly identical SHAP values, 45–57% received different labels (insurance 0.45, wine 0.51, California Housing 0.57 and Bike Sharing 0.51; 3,993 pairs in total), and the share was 0.36 even within the same feature (Fig. 11A). In two wine instances, for example, alcohol had almost the same SHAP value (0.170 and 0.175), but it was labelled standalone in one ($G_4$ = 0.008) and amplified in the other ($G_4$ = 0.078).

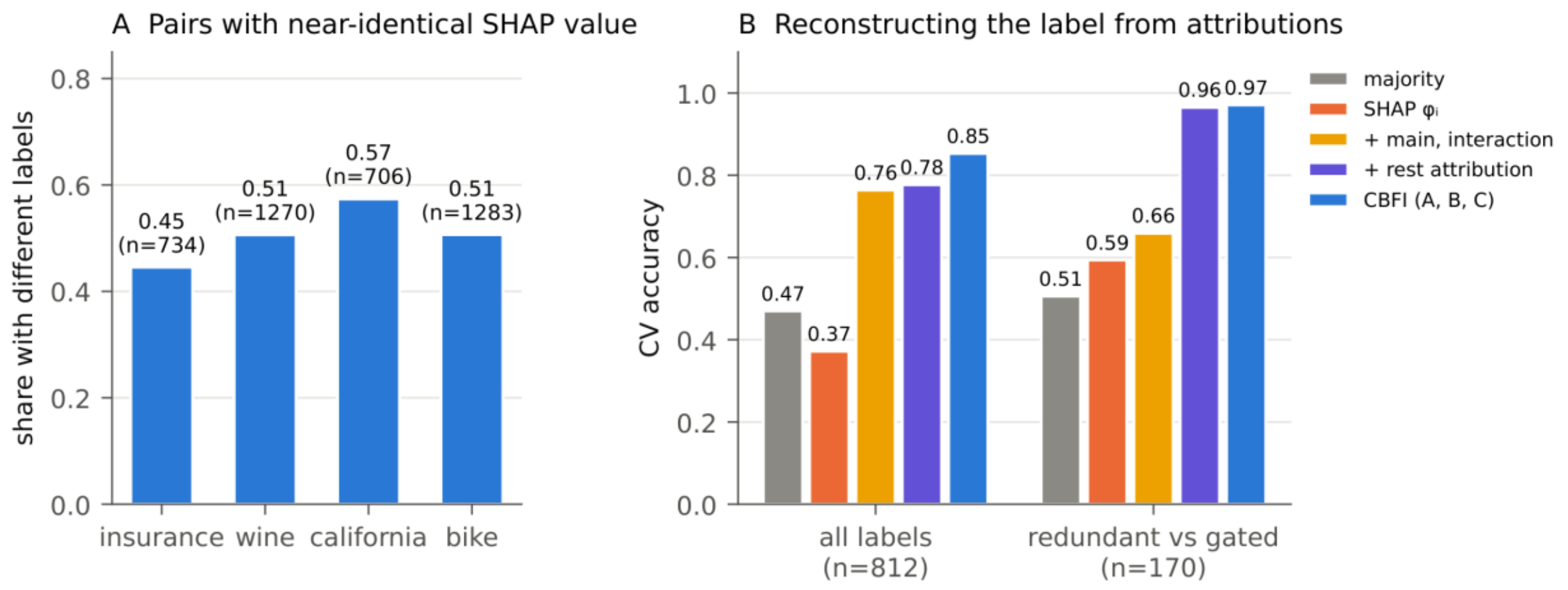

**Figure 11. Information beyond the attribution score.** RF on insurance, wine, California Housing and Bike Sharing; 60 test instances; non-inactive labels with bootstrap confidence ≥ 0.8. (A) Share of feature–instance pairs with nearly identical SHAP values (same sign, within 10%) that receive different CBFI labels. (B) Cross-validated accuracy of predicting the CBFI label from TreeSHAP quantities (gradient boosting, folds grouped by instance), for all labels and for redundant versus gated. "CBFI (A, B, C)" is an upper reference, because the labels are defined from these means.

**Table 13. Predicting the localized CBFI label from TreeSHAP quantities (gradient boosting, 5-fold CV grouped by instance; RF on insurance, wine, california and bike; 60 test instances; non-inactive labels with bootstrap confidence >= 0.8).**

| Labels | n | Inputs | Accuracy | Balanced accuracy |
|---|---|---|---|---|
| **all labels** | 812 | SHAP $\varphi_i$ | 0.372 | 0.211 |
| **all labels** | 812 | + main, interaction | 0.764 | 0.562 |
| **all labels** | 812 | + rest attribution | 0.777 | 0.615 |
| **all labels** | 812 | CBFI (A, B, C) | 0.853 | 0.707 |
| **all labels** | 812 | majority | 0.470 | 0.200 |
| **redundant vs gated** | 170 | SHAP $\varphi_i$ | 0.594 | 0.594 |
| **redundant vs gated** | 170 | + main, interaction | 0.659 | 0.659 |
| **redundant vs gated** | 170 | + rest attribution | 0.965 | 0.965 |
| **redundant vs gated** | 170 | CBFI (A, B, C) | 0.971 | 0.971 |
| **redundant vs gated** | 170 | majority | 0.506 | 0.500 |

*Notes:* $\varphi_i$, SHAP value of the feature; main and interaction: TreeSHAP main effect and summed pairwise interaction values; rest attribution: summed SHAP values of all other features.

When the label was reconstructed from attributions (Fig. 11B, Table 13), the SHAP value alone predicted it worse than the majority class (cross-validated accuracy 0.37 versus 0.47). Adding the TreeSHAP main and interaction parts raised the accuracy to 0.76, and adding the attribution of the remaining features raised it to 0.78. CBFI's own means $(\overline{A}, \overline{B}, \overline{C})$ reached 0.85; this is an upper reference rather than a competitor, because the labels are defined from these means, and it stays below 1 because the labels are bootstrap modes with interval guards. For the distinction between redundant and gated, the SHAP value and its main and interaction parts were close to chance (0.59 and 0.66), whereas adding the attribution of the remaining features raised the accuracy to 0.96. The label therefore adds information to an attribution score, because equal scores often hide different ways of acting. Much of this information can be recovered from a full set of Shapley quantities, but only when they are combined as the taxonomy prescribes, and the remaining-feature component is indispensable for separating substitution from gating.

## Discussion

### *What the method adds to the explanation of a prediction*

The central finding of this study concerns what localized CBFI reports rather than how accurately it detects interactions. For every feature of an instance, the method gives the in-context importance, its split into a standalone part and a context-dependent part, and a label stating how the context acts on the feature. In the insurance model, for example, a conventional attribution reports that smoking raises the predicted charge by a large amount. The decomposition adds two facts to this statement. Most of the effect is standalone (for the smokers among the 50 analysed instances, the mean standalone effect is about 18,800 USD), and the remainder depends on the body-mass index, amplifying the effect for obese smokers and attenuating it for the others. The label condenses this into a statement that a domain expert can check: for this person, the model amplifies the effect of smoking in combination with their BMI.

Interaction values carry part of this information, but not in a form that states the mechanism. The OR/AND example and the mechanism-identification experiment show that substitution and gating produce interaction values of the same sign and size. They become separable once the effect of the remaining features is taken into account, which is exactly what the taxonomy does. This ingredient can also be obtained from Shapley attributions. Applied to k-SII together with the attribution of the remaining features, the taxonomy identified the mechanisms as accurately as localized CBFI, and on real data the labels could be reconstructed from Shapley quantities only when that attribution was included.

We therefore regard the taxonomy as the main conceptual contribution of this study, namely a qualitative layer of explanation on top of attribution scores. Localized CBFI is a natural, model-agnostic way to compute it, because its three quantities are defined directly by hybrid model evaluations at the instance, come with bootstrap confidence, and require no enumeration of pairs or coalitions. Pairwise indices remain the tool of choice for identifying the partner of a feature, and the two approaches are complementary.

### *Accuracy, cost and the choice of method*

On the true functions, localized CBFI and k-SII were equally exact. On fitted models, the dedicated indices recovered interacting pairs slightly more precisely, and exact TreeSHAP interaction values were one to two orders of magnitude faster on tree ensembles. Localized CBFI is also slower than permutation SHAP for full explanations. We therefore see it as a complement to these methods rather than a replacement, suited to explaining selected predictions in detail. When the goal is to measure interaction strength in a tree ensemble, TreeSHAP-IQ is the better choice, and when the goal is to find which pairs interact in an arbitrary model, k-SII is slightly more precise. When a user needs to know how each feature acts in a particular prediction and which part of its importance depends on the context, localized CBFI provides this directly. Two further properties are of practical relevance: the explanations were more stable than those of k-SII under resampled background data, and the single-feature screen identified all interacting features in every scalability setting, which makes a two-stage analysis feasible in higher dimensions.

***Interpreting the decomposition responsibly***

Several interpretations of CBFI in the earlier version of this work needed correction, and these corrections illustrate how the decomposition should be read. First, a standalone effect close to zero does not by itself mean that the feature has no effect on its own. The near-zero standalone effect of smoking reported earlier was an artefact of a routing rule that credited a standalone effect only when the feature alone matched the accuracy of the full model. With the present definition, the standalone effect of smoking is large, and the in-context importance agrees with a counterfactual check.

Second, agreement with SHAP or LIME is no evidence of correctness, and disagreement is no evidence of superiority. Different methods answer different questions with different reference distributions (Krishna et al., 2024). We therefore validated the method against known ground truth and used agreement only descriptively. Third, the decomposition describes the model, not the world. The smoker × BMI pattern is a property of the fitted model; that it matches the obesity–smoking interaction commonly modelled for this dataset (Lantz, 2013) is reassuring, but causal conclusions about medical costs would require a causal analysis. Finally, the labels depend on the relevance threshold $\tau$, and reporting $\tau$ with the explanation, together with the bootstrap confidence of each label, makes this dependence transparent.

***Distinguishing data structure from model structure***

The two-level synthetic design made visible how often fitted models deviate from the generating logic. Random forests introduced small spurious interactions for additive features, and when two features were near-duplicates, the models relied mainly on one of them, so that the duplicate was labelled inactive rather than redundant. An explanation method that is exact on the true function will faithfully report such model behaviour, which is desirable when the goal is to audit the model. Users interested in the data-generating process should nevertheless be aware that explanations of a fitted model can differ from it, and that the difference reflects the model rather than the explanation method.

## Limitations

This study has several limitations. The first concerns cost. Explaining all features of an instance requires about $3np$ model evaluations, which was 1.4–15 times slower than permutation SHAP and much slower than exact TreeSHAP on tree ensembles. The method is therefore suited to detailed explanations of selected predictions rather than to real-time explanation of every prediction. Moreover, the effects are Monte Carlo estimates. With $n = 100$, the variability of the interaction effect is about $8\%$ of the importance and decreases as $1/\sqrt{n}$, but unlike TreeSHAP for tree ensembles, the method offers no exact computation. On fitted models, the recovery of interacting pairs was also slightly less precise than with k-SII and TreeSHAP interaction values.

The second group of limitations concerns the choices that the user must make. Marginal replacement can create unrealistic feature combinations for correlated features, and switching to conditional replacement changed the feature rankings moderately (Kendall $\tau$ 0.47–0.82). The choice should match the purpose of the explanation, and both variants should be reported for strongly correlated data. The labels also depend on the practical threshold $\tau$, which is a user choice, and reversed cases were rare in our benchmarks and identified less reliably (recall 0.80).

Finally, the scope of the taxonomy and of the evaluation is limited. The taxonomy summarizes the interaction between a feature and all remaining features, so higher-order structure is reflected in the labels but not resolved into specific feature sets. The mechanism labels are not exclusive to localized CBFI either: once the taxonomy is defined, comparable labels can be derived from Shapley interaction indices, so the contribution lies in the taxonomy and in a direct estimator for it rather than in information that other methods cannot access. The evaluation covers tabular data with up to 226 features and synthetic functions with known structure, and we did not conduct a user study, so whether the labels help human decision makers remains to be tested.

## Conclusions

In this study, we localized case-based feature importance to individual predictions. The resulting method decomposes the in-context importance of each feature into a standalone and an interaction component and labels how the feature acts: inactive, standalone, amplified, redundant, gated or reversed.

The evaluation against known ground truth and against interaction-specific indices supports three conclusions. First, the decomposition is exact on true functions and comparable, though not superior, to dedicated interaction indices on fitted models. Second, the case taxonomy provides information that attribution scores and interaction values alone do not provide. Equal SHAP values often correspond to different labels, the taxonomy distinguishes substitution from gating without knowing the interacting partner, and its labels agree with the generating logic in 95% of cases on fitted models. Because the same taxonomy can be layered on Shapley interaction indices, it can serve as a general explanation layer. Third, on real data, the decomposition yields explanations that a domain expert can check, such as the BMI-dependent interaction of smoking in medical charges, as well as domain-level diagnostics of how strongly a model relies on context. Localized CBFI is therefore best used alongside established attributions and interaction indices, when a user needs to understand not only which features matter for a prediction but also how they act.

Several directions follow from the limitations above. First, a user study should test whether the labels improve how people anticipate model behaviour and answer counterfactual questions, compared with attribution scores alone. Second, the cost and Monte Carlo variability could be reduced by exact computation of the four hybrid states for tree ensembles, in the spirit of TreeSHAP, and by adaptive sampling that stops once the label is stable. Third, the taxonomy could be extended from a feature and all remaining features to small feature sets, so that the partners responsible for gating or redundancy are identified directly. Finally, the relation between the CBFI estimates and Shapley interaction indices could be formalized, so that the taxonomy can serve as a common explanation layer over different attribution methods.